\documentclass{article}

\usepackage[preprint]{arxiv}

\usepackage[utf8]{inputenc} 
\usepackage[T1]{fontenc}    
\usepackage{hyperref}       
\usepackage{url}            
\usepackage{booktabs}       
\usepackage{amsfonts}       
\usepackage{nicefrac}       
\usepackage{microtype}      

\usepackage{amsmath} 
\usepackage{amssymb} 
\usepackage{blindtext} 
\usepackage[capitalize,nameinlink]{cleveref}[0.19] 
\usepackage{comment} 
\usepackage[pdftex]{graphicx} 
\usepackage{makecell} 
\usepackage{multirow} 
\usepackage{subcaption} 
\usepackage{xfrac} 

\ifpdf
\hypersetup{
  pdftitle={Data augmentation instead of explicit regularization},
  pdfauthor={Alex Hernandez-Garcia, Peter K{\"o}nig},
  pdfkeywords={data augmentation, regularization, image object recognition, neural networks}
}
\fi

\newcommand{\imgpath}{./img}

\title{Data augmentation instead of explicit regularization}

\author{%
  Alex~Hernandez-Garcia\thanks{Corresponding author. Contact: \href{https://alexhernandezgarcia.github.io/}{alexhernandezgarcia.github.io}.\newline~Code available at: \href{https://github.com/alexhernandezgarcia/data-aug-invariance/}{github.com/alexhernandezgarcia/data-aug-invariance}} \\
  Institute of Cognitive Science\\
  University of Osnabr{\"u}ck\\
  Germany\\
  \texttt{ahernandez@uos.de} \\
  \And
  Peter~K{\"o}nig\\
  Institute of Cognitive Science\\
  University of Osnabr{\"u}ck\\
  Germany\\
  \texttt{pkoenig@uos.de} \\
}

\begin{document}

\maketitle

\begin{abstract}
Contrary to most machine learning models, modern deep artificial neural networks typically include multiple components that contribute to regularization. Despite the fact that some (explicit) regularization techniques, such as weight decay and dropout, require costly fine-tuning of sensitive hyperparameters, the interplay between them and other elements that provide implicit regularization is not well understood yet. Shedding light upon these interactions is key to efficiently using computational resources and may contribute to solving the puzzle of generalization in deep learning. Here, we first provide formal definitions of explicit and implicit regularization that help understand essential differences between techniques. Second, we contrast data augmentation with weight decay and dropout. Our results show that visual object categorization models trained with data augmentation alone achieve the same performance or higher than models trained also with weight decay and dropout, as is common practice. We conclude that the contribution on generalization of weight decay and dropout is not only superfluous when sufficient implicit regularization is provided, but also such techniques can dramatically deteriorate the performance if the hyperparameters are not carefully tuned for the architecture and data set. In contrast, data augmentation systematically provides large generalization gains and does not require hyperparameter re-tuning. In view of our results, we suggest to optimize neural networks without weight decay and dropout to save computational resources, hence carbon emissions, and focus more on data augmentation and other inductive biases to improve performance and robustness.
\end{abstract}


\section{Introduction}
\label{sec:intro}

One of the central issues in machine learning is finding ways of improving generalization. Regularization, loosely defined as any modification applied to a learning algorithm that helps the model generalize better, plays therefore a key role in machine learning. In the case of deep learning, where neural networks tend to have several orders of magnitude more parameters than training examples, statistical learning theory \cite{vapnik1971vc} indicates that regularization becomes even more crucial. Accordingly, a myriad of techniques have been proposed as regularizers: weight decay \cite{hanson1989wd} and other $L^p$ penalties on the learnable parameters; dropout---random dropping of units during training---\cite{srivastava2014dropout} and stochastic depth---random dropping of whole layers---\cite{huang2016stochasticdepth}, to name a few. 

Moreover, whereas in simpler machine learning algorithms the regularizers can be easily identified as explicit terms in the objective function, in modern deep neural networks the sources of regularization are not only explicit but implicit \cite{neyshabur2014implicitreg}. In this regard, many techniques have been studied for their regularization effect, despite not being explicitly intended as such. Examples are convolutional layers \cite{lecun1990conv}, batch normalization \cite{ioffe2015batchnorm} and data augmentation---synthetically creating new examples through transformations of the existing data points. In sum, there are multiple elements in deep learning that provide inductive biases and thus contribute to improving generalization.

It is common practice in both the scientific literature and application to incorporate several of these regularization techniques in the training procedure of neural networks. For instance, weight decay, dropout and data augmentation have been used jointly in multiple well-known architectures \cite{tan2019efficientnet, huang2017densenet, zagoruyko2016wrn, springenberg2014allcnn}. It is therefore implicitly assumed that each technique is necessary and contributes additively to improving generalization. However, the interplay between regularization techniques is yet to be well understood and might be an important piece for the puzzle of why and when deep networks generalize.

Understanding the interaction and contribution of regularization techniques is not only interesting from a theoretical point of view, but it also has an impact on the computational resources used to train deep neural networks, as there exist decisive differences across regularization methods. For example, explicit regularization techniques such as weight decay and dropout, heavily depend on hyperparameter tuning. Moreover, they regularize the model by constraining the representational capacity. Ironically, a trend has been to find ways of training deeper and wider networks with larger capacity \cite{simonyan2014, he2016resnet, zagoruyko2016wrn, canziani2016analysisdl}, which is eventually reduced through regularization. It is known, for instance, that the gain in generalization provided by dropout comes at the cost of using larger models and training for longer \cite{goodfellow2016dlbook}. In contrast, data augmentation neither reduces the representational capacity nor depends on sensitive hyperparameters. Despite its popularity, data augmentation is often looked down by the neural networks community as a method that should not be used to assess the potential of proposed architectures or methods. This work seeks to rethink data augmentation and analyze some of its advantages, in contrast to the most popular regularization techniques.

Our specific contributions are the following:

\begin{itemize}
 \item Propose definitions of explicit and implicit regularization that aim at solving the ambiguity in the literature (\cref{sec:expl_impl_reg}).
 \item A theoretical discussion based on statistical learning theory about the differences between explicit regularization and data augmentation, which highlight the advantages of the latter (\cref{sec:theoretical}).
 \item An empirical analysis of the performance of models trained with and without explicit regularization, and different levels of data augmentation on several benchmarks (methodology in \cref{sec:methods} and results in \cref{sec:results}). Further, we study their adaptability to learning from fewer examples (\cref{sec:less_data}) and to changes in the architecture (\cref{sec:depth}).
 \item A discussion on why encouraging data augmentation instead of explicit regularization can benefit both theory and practice in deep learning (\cref{sec:discussion}).
\end{itemize}

\section{Explicit and implicit regularization}
\label{sec:expl_impl_reg}

While several regularization taxonomies have been proposed, to the best of our knowledge there is no formal definitions of explicit and implicit regularization in the machine literature. Nonetheless, the terms have been widely used \cite{neyshabur2014implicitreg, zhang2016understandingdl, wilson2017neurips, mesnil2011transferlearning, poggio2017theory3, martin2018selfregularisation, achille2018emergence}. This could suggest that the concepts are ingrained in the field and well understood by the community. However, by analyzing the use of the terms explicit and implicit regularization in the literature one can see that there is a high degree of ambiguity. Before providing the definitions, we briefly review some examples and motivate the need for formal definitions.

The PhD thesis by Behnam Neyshabur \cite{neyshabur2017thesis} was devoted to the study of implicit regularization in deep learning. For instance, Neyshabur showed that common optimization methods for deep learning, such as stochastic gradient descent (SGD), introduce an inductive bias that lead to better generalization. That is, SGD \textit{implicitly} regularizes the learning process. However, the notion and definition of implicit regularization is only vaguely implied in Neyshabur's PhD thesis and related works, as the generalization improvement provided by techniques that are not \textit{typically} considered as regularization, such as SGD. By extension, explicit regularization would refer to those other techniques: ``we are not including any explicit regularization, neither as an explicit penalty term nor by modifying optimization through, e.g., drop-outs, weight decay, or with one-pass stochastic methods'' \cite{neyshabur2017thesis}.

In \cite{poggio2017theory3}, it can be interpreted that implicit regularization refers to techniques that lead to minimization of the parameter norm without explicitly optimizing for it. By extension, explicit regularization would refer to classical penalties on the parameter norm, such as weight decay. It is therefore unclear whether other methods such as dropout should be considered explicit or implicit regularization according to \cite{poggio2017theory3}.

\citet{zhang2016understandingdl} raised the thought-provoking idea that ``explicit regularization may improve generalization performance, but is neither necessary nor by itself sufficient for controlling generalization error.'' The authors came to this conclusion from the observation that turning off the ``explicit regularizers'' of a model does not prevent the model from generalizing reasonably well. In their experiments, the explicit regularization techniques they turned off were, specifically, weight decay, dropout and data augmentation. In this case, it seems that \citet{zhang2016understandingdl} made a distinction based on the mere intention of the practitioner. Under that logic, because data augmentation has to be designed and applied \textit{explicitly}, it would be explicit regularization. 

These examples illustrate that the terms explicit and implicit regularization have been used subjectively and inconsistently in the literature. In order to help avoid ambiguity, settle the concepts and facilitate future discussion about explicit and implicit regularization, in the next section we propose definitions and provide examples to illustrate each category.

\subsection{Definitions and examples}
\label{sec:reg-definitions}

We propose the following definitions of explicit and implicit regularization:

\begin{itemize}
\item \textbf{Explicit regularization techniques} are those techniques which reduce the \textit{representational} capacity of the model class they are applied on. That is, given a model class $\mathcal{H}_0$, for instance a neural network architecture, the introduction of explicit regularization will span a new hypothesis set $\mathcal{H}_1$,  which is a \textit{proper subset} of the original set, that is $\mathcal{H}_1 \subsetneq \mathcal{H}_0$.
\item \textbf{Implicit regularization} is the reduction of the generalization error or overfitting provided by means other than explicit regularization techniques. Elements that provide implicit regularization do not reduce the \textit{representational} capacity, but may affect the \textit{effective} capacity of the model: the \textit{achievable} set of hypotheses given the model, the optimization algorithm, hyperparameters, etc.
\end{itemize}

Note that we define explicit and implicit regularization by using the concepts of \textit{representational} and \textit{effective} capacity. Although these terms are also used ambiguously by some practitioners, definitions of these concepts can be found in the literature. For instance, the textbook Deep Learning \cite{goodfellow2016dlbook} defines the representational capacity as the ``the family of functions the learning algorithm can choose from'' and explains that the effective capacity ``may be less than the representational capacity'' because the learning algorithm does not always find the ``best function'' due to ``limitations, such as the imperfection of the optimization algorithm''. We here adopt these definitions of representational and effective capacity.

One of the most common explicit regularization techniques in machine learning is $L^p$-norm regularization \cite{tikhonov1963regularisation}, of which weight decay is a particular case, widely used in deep learning. Weight decay sets a penalty on the $L^2$ norm of the model's learnable parameters, thus constraining the representational capacity of the model. Dropout \cite{srivastava2014dropout} is another common example of explicit regularization, where the hypothesis set is reduced by stochastically deactivating a number of neurons during training. Similar to dropout, stochastic depth \cite{huang2016stochasticdepth}, which drops whole layers instead of neurons, is also an explicit regularization technique.

Regarding implicit regularization, note first that the above definition does not refer to \textit{techniques}---as in the definition of explicit regularization---but to a regularization \textit{effect}, as it can be provided by multiple elements of different nature. For instance, stochastic gradient descent (SGD) is known to have an implicit regularization effect---reduction of the generalization error---without constraining the representational capacity \cite{zhang2017sgd}. Batch normalization neither reduces the capacity, but it improves generalization by smoothing the optimization landscape \cite{santurkar2018bn}. Of quite a different nature, but still implicit, is the regularization effect provided by early stopping \cite{yao2007earlystopping}, which does not reduce the representational but the effective capacity.

In these examples and all other cases of implicit regularization, we can think of the effect on the capacity in the following way: we start by defining our model class, for instance a neural network, which spans a set of functions $\mathcal{H}_0$. If we decide to train with explicit regularization, for instance weight decay or dropout, then the model will have access to a smaller set of functions $\mathcal{H}_1 \subsetneq \mathcal{H}_0$, that is smaller representational capacity. On the contrary, if we decide to train with SGD, batch normalization or early stopping, the set of functions spanned by the model stays identical. Due to the dynamics and limitations imposed by these techniques, some functions may never be found, but theoretically they could be. In other words, the effective capacity may be smaller but not the representational capacity.

Central to this paper is data augmentation, a technique that provides an implicit regularization effect. As we have discussed, \citet{zhang2016understandingdl} considered data augmentation an explicit regularization technique, in the same category as weight decay and dropout. However, data augmentation does not reduce the representational capacity of the models and hence, according to our definitions, cannot be considered explicit regularization. This is relevant to understand the differences between weight decay, dropout and data augmentation that we present in the rest of this paper.

\section{Theoretical insights}
\label{sec:theoretical}
The hypothesis that we aim to prove in this paper is that the generalization gain provided by explicit regularization, particularly weight decay and dropout, is overshadowed by the inductive biases typically incorporated into neural network training, particularly through data augmentation. While in \cref{sec:methods,sec:results} we provide empirical evidence, in this section we provide some insights about the effect on generalization of these techniques that support the hypothesis, from the point of view of statistical learning theory.

The generalization of a model class $\mathcal{H}$ can be analyzed through complexity measures such as the VC-dimension or, more generally, the Rademacher complexity $\mathcal{R}_{N}(\mathcal{H}) = \mathbb{E} \left[ \hat{\mathcal{R}}_{N}(\mathcal{H}) \right]$, where:
\begin{equation}
\label{eq:daugreg-rademacher}
  \hat{\mathcal{R}}_{N}(\mathcal{H}) = \mathbb{E}_{\sigma} \left[ \underset{h \in \mathcal{H}}{\mathrm{sup}} \left| \frac{1}{N} \sum_{i=1}^{N} \sigma_{i}h(x_{i}) \right| \right]
\end{equation}
is the empirical Rademacher complexity, defined with respect to a specific set of $N$ data samples. Then, in the case of binary classification and the class of linear separators, the generalization error of a hypothesis, $\hat{\epsilon}_{N}(h)$, can be bounded using the Rademacher complexity:

\begin{equation}
\label{eq:daugreg-genbound}
 \hat{\epsilon}_{N}(h) \leq \mathcal{R}_{N}(\mathcal{H}) + \mathcal{O} \left( \sqrt{\frac{\ln \sfrac{1}{\delta}}{N}} \right)
\end{equation}
with probability $1 - \delta$. Tighter bounds for some model classes, such as fully connected neural networks, can be obtained \cite{bartlett2002rademacher}, but it is not trivial to formally analyze the influence on generalization of specific architectures or techniques. Nonetheless, we can use these theoretical insights to discuss the differences between weight decay, dropout and data augmentation. 

First, a straightforward yet very relevant conclusion from the analysis of any generalization bound is the strong dependence on the number of training examples $N$. Increasing $N$ drastically improves the generalization guarantees, as reflected by the second term in the right hand side of Equation~\ref{eq:daugreg-genbound} and by the dependence of the Rademacher complexity (Equation~\ref{eq:daugreg-rademacher}) on the sample size too. In most analyses, the number of examples is consider a given constant---for instance, the size of a benchmark data set---over which we have no control. However, \textit{perceptually plausible}\footnote{In this work, we consider exclusively \textit{perceptually plausible} image data augmentation, which we define as the transformations that preserve the properties of the perceived visual world of humans as well as, in the case of image object recognition, the object class. Other types of transformations in which perceptual plausibility is not preserved have been successfully used in computer vision. One popular example is \textit{mixup} \cite{zhang2017mixup}, which performs the weighted average of the pixels of two images---and their labels. Another example is data augmentation in feature space \cite{devries2017daugfeatspace}} data augmentation exploits prior knowledge about the data domain and aspects of visual perception---in the case of image object recognition---to create new examples and its impact on generalization is related to an increment in $N$, as stochastic data augmentation can generate virtually infinite different samples. Admittedly, the augmented samples are not independent and identically distributed and thus, the effective increment of samples does not strictly correspond to the increment in $N$. This is why formally analyzing the impact of data augmentation on generalization is complex and currently an open question. Recent studies have made progress in this direction by analyzing the effect of simple data transformations on generalization from a theoretical point of view \cite{chen2019invariance, rajput2019daug}.

In contrast, explicit regularization methods aim at improving the generalization error by constraining the hypothesis class $\mathcal{H}$ to reduce its complexity $\mathcal{R}_{N}(\mathcal{H})$ and, in turn, the generalization error $\hat{\epsilon}_{N}(h)$. Crucially, while data augmentation exploits domain knowledge, most explicit regularization methods only \textit{naively} constrain the hypothesis class, by simply reducing the representational capacity, as we have discussed in \cref{sec:reg-definitions}. For instance, weight decay constrains the learnable functions $\mathcal{H}$ by setting a penalty on the weights norm. Interestingly, \citet{bartlett2017boundsnn} showed that weight decay has little impact on the generalization bounds and confidence margins. Dropout has been extensively used and studied as a regularization method for neural networks \cite{wager2013dropout}, but the exact way in which it impacts generalization is still an open question. In fact, it has been stated that the effect of dropout on neural networks is ``somewhat mysterious'', complicated and its penalty highly non-convex \cite{helmbold2017dropout}. Recently, \citet{mou2018dropout} established new generalization bounds on the variance induced by a specific type of dropout on feedforward networks. 

A relevant observation is that the noise introduced by dropout through the random dropping of neural network units can be projected back into the input space \cite{bouthillier2015dropoutasdaug}. That is, dropout can be analyzed as a random form of data augmentation without domain knowledge. This implies that any generalization bound derived for dropout can be regarded as a pessimistic bound for domain-specific, standard data augmentation. A similar argument applies for weight decay, which, as first shown by \citet{bishop1995tikhonov}, is equivalent to training with noisy examples if the noise amplitude is small and the objective is the sum-of-squares error function. Therefore, some forms of explicit regularization are at least approximately equivalent to adding random noise to the training examples, which is the simplest form of data augmentation\footnote{Note that the opposite view---domain-specific data augmentation as explicit regularization---does not apply. In the supplementary material (\cref{sec:suppl-reg_taxonomy}) we discuss the taxonomies of regularization, including the difference between data augmentation and data-dependent regularization}. Thus, it is reasonable to argue that the inductive biases of more sophisticated data augmentation can overshadow the benefits provided by explicit regularization.

In general, we argue that the reason why explicit regularization may not be necessary is that neural networks are already implicitly regularized by many elements---stochastic gradient descent (SGD), convolutional layers, normalization and data augmentation, to name a few---that provide more successful inductive biases \cite{neyshabur2014implicitreg}. For instance, it has been shown that linear models optimized with SGD converge to solutions with small norm, without any explicit regularization \cite{zhang2016understandingdl}. Furthermore, as we will further discuss in \cref{sec:discussion}, if over-parameterized neural networks are able to generalize well, the need for constraining their capacity is questionable.

Building upon these initial theoretical insights, in \cref{sec:methods} we present an empirical study to contrast data augmentation and explicit regularization---weight decay and dropout.

\section{Methods}
\label{sec:methods}

This section describes the main aspects of the experimental setup for systematically analyzing the role of data augmentation in deep neural networks compared to weight decay and dropout. The methods build upon those used in preliminary studies \cite{hergar2018wddropout,zhang2016understandingdl}. Extended details can be consulted in the supplementary material and the code uploaded to \href{https://github.com/alexhernandezgarcia/data-aug-invariance/}{github.com/alexhernandezgarcia/data-aug-invariance}.

\subsection{Data}
\label{sec:data}

We performed the experiments on the highly benchmarked data sets ImageNet \cite{russakovsky2015imagenet} ILSVRC 2012, CIFAR-10 and CIFAR-100 \cite{krizhevsky2009cifar}. We resized the 1.3 M images from ImageNet into $150\times200$ pixels, as a good compromise between keeping a high resolution and speeding up the training \cite{touvron2019resolution}. Both on ImageNet and on CIFAR, the pixel values are in the range $[0, 1]$.  

So as to analyze the role of data augmentation, we trained every model with two different augmentation schemes as well as with no data augmentation at all. The two augmentation schemes are the following:

 \textbf{\textit{Light} augmentation}: This scheme is common in the literature \cite{goodfellow2013maxout, springenberg2014allcnn} and performs only horizontal flips and horizontal and vertical translations of 10\% of the image size. 

 \textbf{\textit{Heavier} augmentation}: This scheme performs a larger range of affine transformations such as scaling, rotations and shear mappings, as well as contrast and brightness adjustment. On ImageNet we additionally perform random crops of $128\times128$ pixels. The choice of the allowed transformations is arbitrary and the only criterion was that the objects be still recognizable in general. We deliberately avoided designing a particularly successful scheme. Further details can be consulted in the supplementary material (\cref{sec:suppl-daug_details}).

\subsection{Network architectures}
\label{sec:archs}

We trained three distinct, popular architectures that have achieved successful results in visual object recognition: the all convolutional network, All-CNN \cite{springenberg2014allcnn}; the wide residual network, WRN \cite{zagoruyko2016wrn}; and the densely connected network, DenseNet \cite{huang2017densenet}. Importantly, we kept the training hyperparameters (learning rate, training epochs, batch size, optimizer, etc.) as in the original papers. Table~\ref{tab:architectures} summarizes the main features of each network and more details can be found in the supplementary material (\cref{sec:suppl-architectures}) and code.

\begin{table}[ht]
\caption{Key aspects of the architectures. Cells with two values correspond to ImageNet~/~CIFAR.}
\begin{center}
\begin{tabular}{rccc}
    & \textbf{All-CNN} & \textbf{WRN} & \textbf{DenseNet}\\
    Ref. in original paper & \textit{All-CNN-C} & \textit{WRN-28-10} & \textit{DenseNet-BC}\\
    Main feature & Only conv. layers & Residual connections & Dense connectivity\\
    Number of layers & 16 / 12 & 28 & 101\\
    Number of parameters & 9.4 / 1.3 M & 36.5 M & 0.8 M\\
    Training hours & 35--45 / 2.5 & 100--145 / 14--15  & 24--27\\
    CO2e emissions\footnotemark (kg) & 4.17--5.36 / 0.29 & 11.91--17.27 / 1.66--1.78  & 2.86--3.21\\
\end{tabular}
\end{center}
\label{tab:architectures}
\end{table}

\footnotetext{The carbon emissions have been computed using the online calculator at \href{http://www.green-algorithms.org/}{green-algorithms.org} \cite{lannelongue2020carbonemissions}. The whole set of experiments emmited an estimated total of 390.45 CO2e. Details about how the carbon emissions were calculated and about the impact of this study on global warming can be found in the supplementary material (\cref{sec:suppl-carbon_footprint}).}

\subsection{Train and test}

Every architecture was trained on each data set both with explicit regularization---weight decay and dropout as specified in the original papers---and without. Furthermore, we trained each model with the three data augmentation schemes: no augmentation, light and heavier. The performance of the models was computed on the held out test set. As in previous works \cite{krizhevsky2012alexnet, simonyan2014}, we averaged the softmax posteriors over 10 random \textit{light} augmentations, since slightly better results are obtained.

All the experiments were performed on Keras \cite{chollet2015keras} on top of TensorFlow \cite{tensorflow2015} and on a single GPU NVIDIA GeForce GTX 1080 Ti.

\begin{figure}[htbp]
  \begin{center}
    \includegraphics[width = \textwidth]{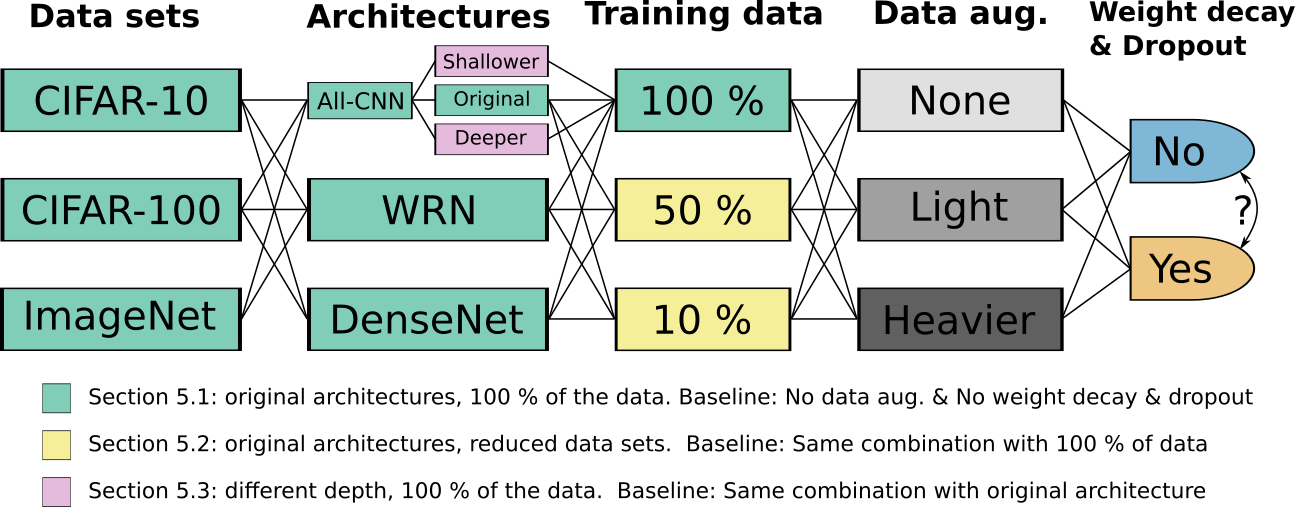}
  \end{center}
  \caption{Visual summary of the experimental setup. The figure represents the factors of variation in our experiments: data sets, architectures, amount of training data, data augmentation scheme and inclusion of explicit regularization. Comparisons within a factor of variation are most relevant on the factors on the right, like the performance of the models train with and without explicit regularization.}
  \label{fig:experimental_setup}
\end{figure}

\section{Results}
\label{sec:results}

In this section, we present the results of the empirical study. In the first set of experiments (\cref{sec:orig_nets}) we trained the architectures as in the original papers with the full data sets. A relevant characteristic of explicit regularization methods is that they typically depend on hyperparameters. These are usually fine-tuned by the authors of research papers to achieve higher performance, as demanded by the dynamics of the scientific publication environment in the machine learning community. However, the sensitivity of the results towards these hyperparameters is often not made available. In order to gain insight on the role of explicit regularization and data augmentation in real world cases, where the hyperparameters have not been highly optimized, we varied the amount of training data (\cref{sec:less_data}) and the depth of the architectures (\cref{sec:depth}), while keeping all other hyperparameters untouched.

The objective of this study is to contrast the performance gained by training the models with both explicit regularization and data augmentation, which is the common practice in the literature \cite{tan2019efficientnet, huang2017densenet, zagoruyko2016wrn, springenberg2014allcnn}, against training with only data augmentation. Hence, the presentation of the results in the figures aims at facilitating this comparison. In the performance plots, we represent the relative performance gain of each model with respect to the relevant baseline, which we specify at each section. We plot the results of each model in pairs: the squared blue dots on the top, blue-shaded area correspond to the models trained with only data augmentation and the round orange dots on the bottom, orange-shaded area to the models trained with both data augmentation and explicit regularization. Additionally, the results of training with different levels of data augmentation are represented with dots in three lightness and saturation shades and we connected with dotted lines the models trained with the same level of augmentation.

In order to assess the statistical significance of the differences between models trained with and without explicit regularization, we carried out percentile bootstrap analyses \cite{efron1992bootstrap}, that is simulations based on sampling with replacement. We followed the guidelines by \citet{rousselet2019bootstrap}. In all cases, the values of the distribution correspond to the difference between the performance---with respect to the baseline---of the models trained without explicit regularization minus the performance of the models trained with explicit regularization. We then compared the distribution of this difference in the bootstrap samples and with respect to the null hypothesis, that is no difference ($H_0 = 0$). For each experiment we sampled all possible bootstrap samples with replacement or a maximum of one million.

\subsection{An alternative to explicit regularization}
\label{sec:orig_nets}

\begin{figure}[htbp]
  \begin{center}
    \includegraphics[width = \textwidth]{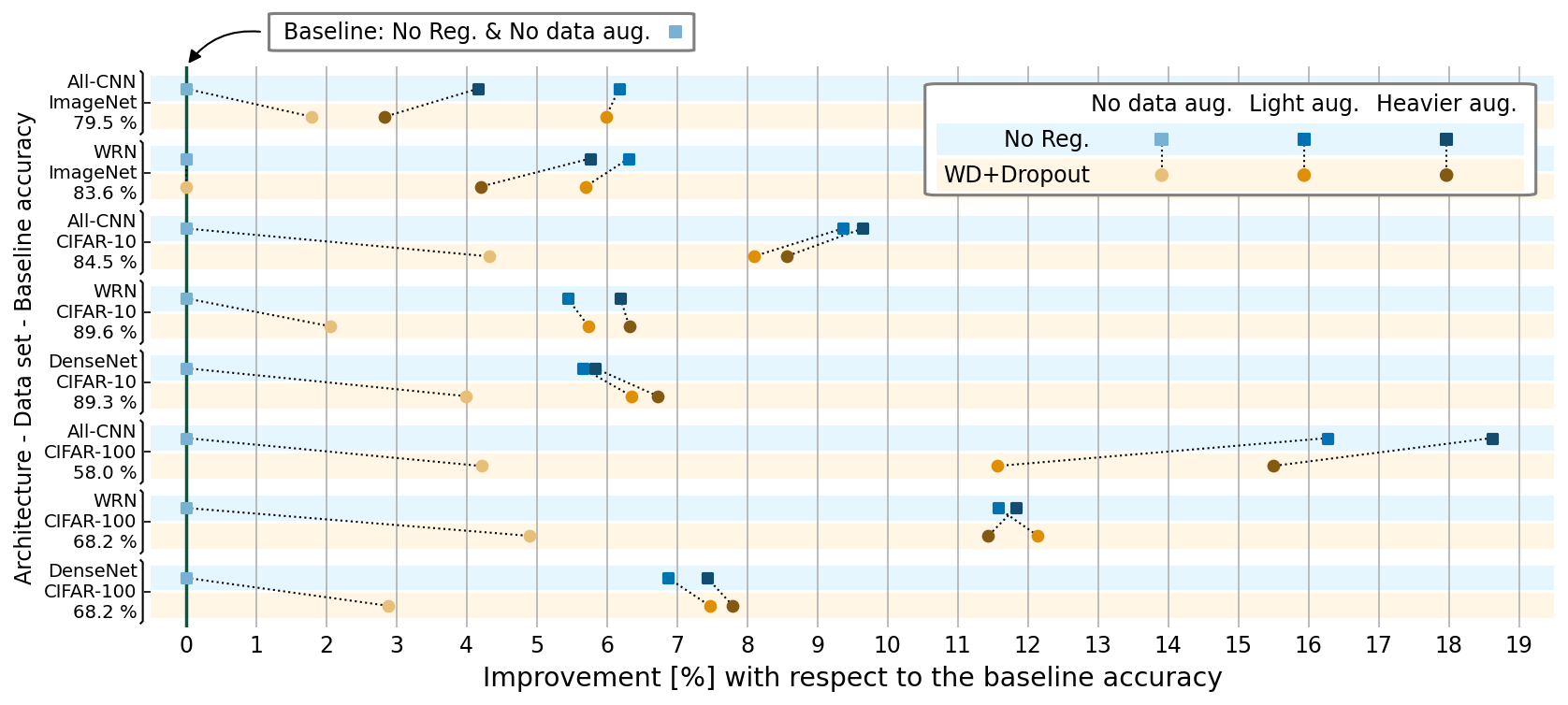}
  \end{center}
  \caption{Relative improvement of adding data augmentation and explicit regularization to the baseline models, $(accuracy - baseline)/accuracy * 100$. The baseline accuracy is shown on the left. The results suggest that data augmentation alone (in blue) can achieve even better performance than the models trained with both weight decay and dropout (in orange).}
  \label{fig:orig}
\end{figure}

\begin{figure}[htbp]
  \centering
  \begin{center}
    \includegraphics[width = \textwidth]{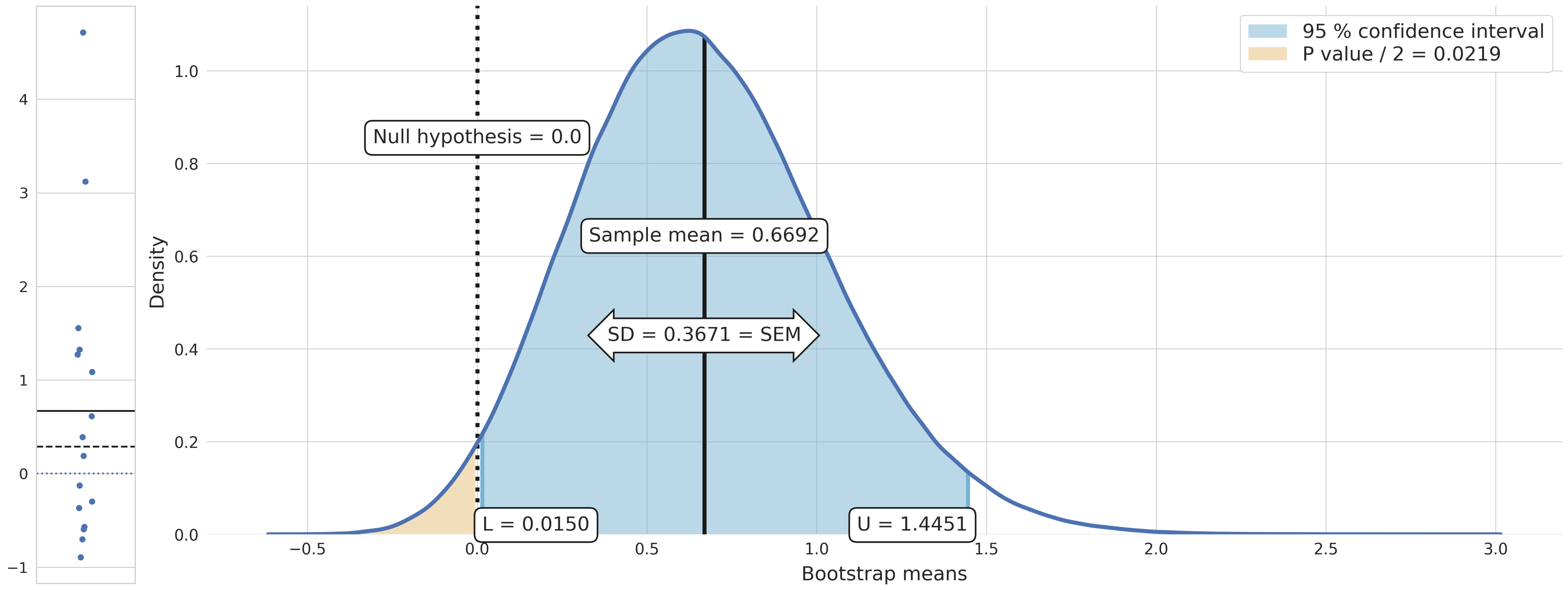}
  \end{center}
  \caption{Bootstrap analysis to assess the difference in performance gain provided by training without and with weight decay and dropout, on the original architectures and using the full data sets. On the left of the figure we plot the bootstrap values---differences---with the mean and median as a solid and dashed line, respectively. The main figure shows the distribution of the mean of the bootstrap samples, the standard error of the sample mean, the 95~\% confidence intervals and the $P$ value with respect to the null hypothesis ($H_0=0$).}
  \label{fig:bootstrap_orig}
\end{figure}

First, we contrast the regularization effect of data augmentation and explicit regularization---weight decay and dropout---on the original networks trained with the complete data sets, and show the results in \cref{fig:orig}. As a baseline, we consider the ``bare bone'' models, that is the model trained with neither explicit regularization nor data augmentation. We report the accuracy of the baseline model on the left axis of the plot in \cref{fig:orig}. So as to assess the relevant comparisons, we show the relative improvement in test performance achieved by adding each technique or combination of techniques to the baseline model. \Cref{fig:bootstrap_orig} shows the results of the bootstrap analysis, which considers the differences of all pairs---squared blue dots minus round orange dots, connected with dotted lines. \cref{tab:orig_nets} shows the mean and standard deviation of each combination across architectures and data sets\footnote{The relative performance of WRN on ImageNet trained with weight decay and dropout with respect to the baseline is negative (-6.22~\%) and is neither depicted in \cref{fig:orig} nor taken into consideration to compute the average improvements in \cref{tab:orig_nets} and the bootstrap analysis in \cref{fig:bootstrap_orig}.}.

\begin{table}[htbp]
\caption{Average accuracy improvement over the baseline model of each combination of data augmentation level and presence of weight decay and dropout.}
\begin{center}
\begin{tabular}{rcc}
            & No explicit reg.  & Weight decay \& dropout \\
    None    & \textit{baseline} & 3.02 (1.65)            \\
    Light   & 8.46 (3.80)       & 7.88 (2.60)            \\
    Heavier & 8.68 (4.69)       & 7.92 (4.03) 
\end{tabular}
\end{center}
\label{tab:orig_nets}
\end{table}

The first conclusion from \cref{fig:orig,fig:bootstrap_orig} as well as \cref{tab:orig_nets} is that training with data augmentation alone (blue dots on the top, blue-shaded areas) achieves better accuracy than training with both augmentation and explicit regularization (in orange). This is the case in more than half of the cases (9/16) and the bootstrap analysis reveals that the difference is positive with 95~\% confidence and $P$ value $=0.0220$. On average, adding data augmentation to the baseline model improved the accuracy on 8.57~\%, and adding both augmentation and explicit regularization on 7.90~\%.

At first glance, one may think that this is not remarkable, since the differences are small and data augmentation alone is not better in 100~\% of the cases. However, this result is surprising and remarkable for the following reason: note that the studied architectures achieved state-of-the-art results at the moment of their publication and the models included all light augmentation, weight decay and dropout, whose parameters were presumably finely tuned to optimize the accuracy. The replication of these results corresponds to the mid-orange dots in \cref{fig:orig}. Here, we have shown that simply removing weight decay and dropout---while keeping all other hyperparameters intact, see \cref{sec:archs}---improves the \textit{then state-of-the-art} accuracy in 4 of the 8 studied cases. Why did not the authors trained without explicit regularization and obtain better results?

Second, it can also be observed that the regularization effect of weight decay and dropout, an average improvement of 3.02~\% with respect to the baseline, is much smaller than that of data augmentation: simply applying light augmentation increased the accuracy in 8.46~\% on average. Although the heavier augmentation scheme was deliberately not designed to optimize the performance, in both CIFAR-10 and CIFAR-100 it improved the test performance with respect to the light augmentation scheme. This was not the case on ImageNet, probably due to the larger complexity of the data set, among other factors.

\begin{figure}[htbp]
  \begin{center}
    \includegraphics[width = \textwidth]{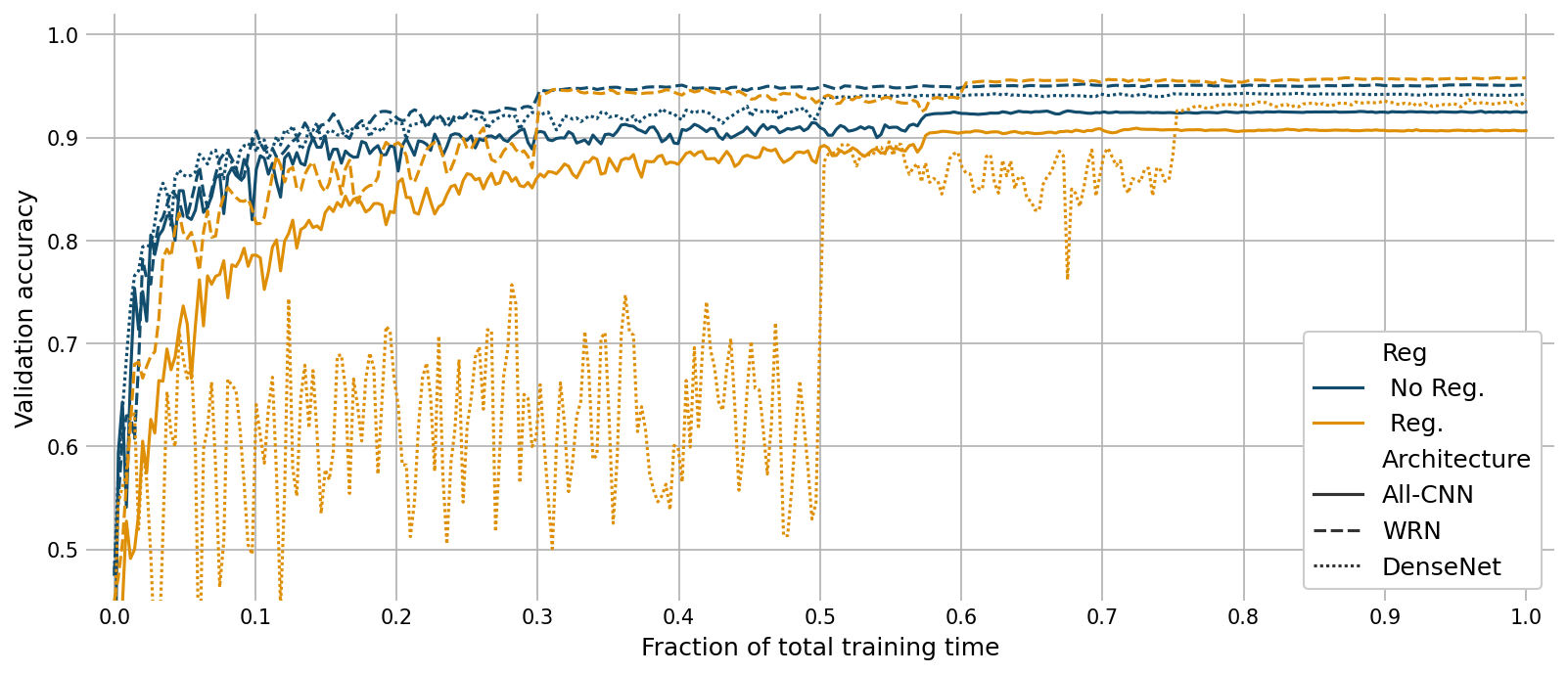}
  \end{center}
  \caption{Dynamics of the validation accuracy during training of All-CNN, WRN and DenseNet, trained on CIFAR-10 with heavier data augmentation, contrasting the models trained with explicit regularization (orange lines) and the models trained with only data augmentation (in blue). The regularized models heavily rely on the learning rate decay to obtain the boost of performance, while the models trained without explicit regularization quickly approach the final performance.}
  \label{fig:dynamics}
\end{figure}

Further, it can be observed that the results are in general more consistent in the models trained without explicit regularization. Finally, an additional advantage of training without explicit regularization is that the learning dynamics (\cref{fig:dynamics}) is much faster and predictive of the final performance. Typically, regularizers such as weight decay and dropout effectively prevent the model from fitting the training data during the first epochs and heavily rely on the learning rate decay to obtain the boost that yields the final performance. This is particularly acute on DenseNet, which performs heavier weight decay. On the contrary, models trained with only data augmentation reach very high validation performance after a few epochs and have less dependence on the learning rate schedule.

In sum, the performance gain provided by weight decay and dropout can be achieved and often improved by data augmentation alone. Besides, the models trained without explicit regularization presented additional advantages, which we further discuss in \cref{sec:discussion}.

\subsection{Fewer available training examples}
\label{sec:less_data}

\begin{figure}[htbp]
  \centering
      \subfloat[50~\% of the available training data]{%
          \label{fig:less_data_50}%
          \includegraphics[keepaspectratio=true, width=\linewidth]{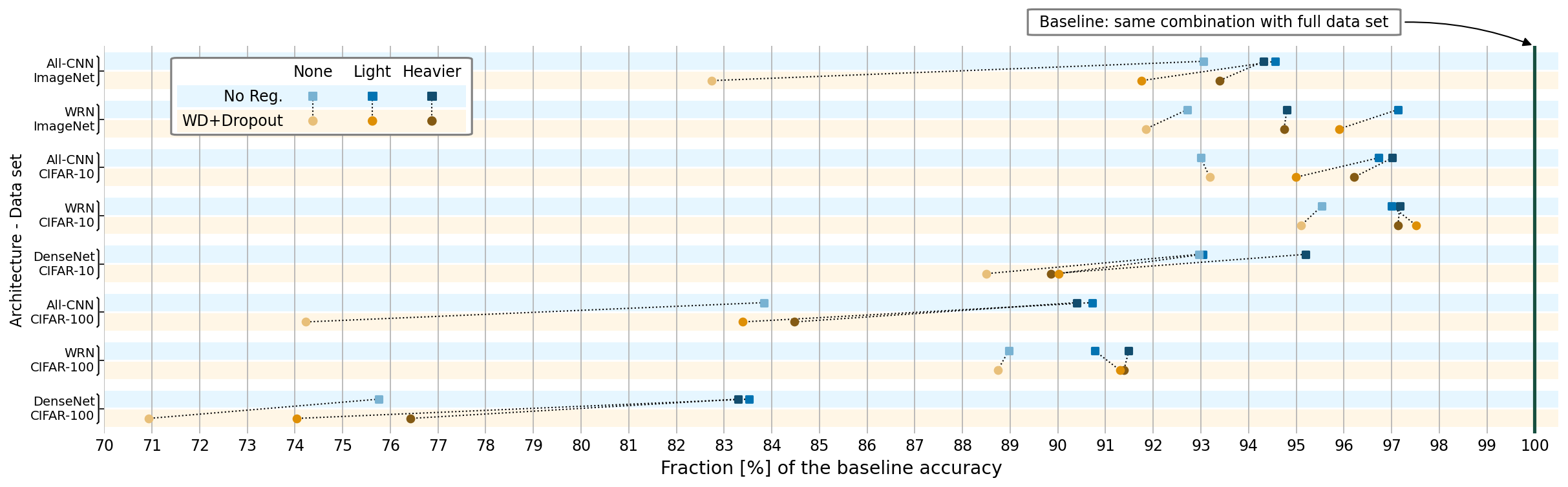}%
      }

      \subfloat[10~\% of the available training data]{%
          \label{fig:less_data_10}%
          \includegraphics[keepaspectratio=true, width=\linewidth]{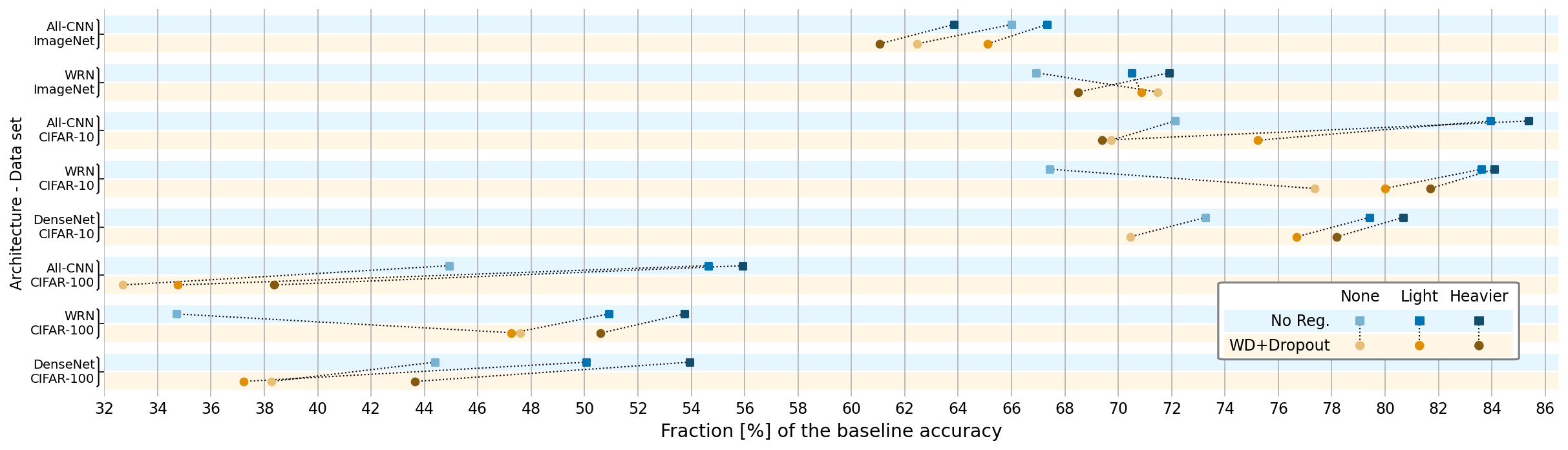}%
      }
  \caption{Fraction of the baseline performance when the amount of available training data is reduced, $accuracy/baseline * 100$. The models trained with explicit regularization present a significant drop in performance as compared to the models trained with only data augmentation. The differences become larger as the amount of training data decreases.}
  \label{fig:less_data}
\end{figure}

\begin{table}[htbp]
  \caption{Average fraction of the original accuracy of each corresponding combination of data augmentation level and presence of weight decay and dropout.}
  \begin{center}
    \begin{tabular}{rcc}
      & \multicolumn{2}{c}{50~\% of the training data}    \\
      \cline{2-3} 
              & No explicit reg. & Weight decay \& dropout \\
      None    & 88.11 (6.27)     & 83.20 (9.83)           \\
      Light   & 91.47 (4.31)     & 88.27 (7.39)           \\
      Heavier & 91.82 (4.63)     & 89.28 (6.63)           \vspace{3pt} \\
      
      & \multicolumn{2}{c}{10~\% of the training data}    \\
      \cline{2-3} 
              & No explicit reg. & Weight decay \& dropout \\
      None    & 58.72 (14.93)    & 58.75 (16.92)          \\
      Light   & 67.55 (14.27)    & 60.89 (18.39)          \\
      Heavier & 68.69 (13.61)    & 61.43 (15.90) 
    \end{tabular}
  \end{center}
  \label{tab:less_data}
\end{table}

\begin{figure}[htbp]
  \centering
      \subfloat[50~\% of the available training data]{%
		  \label{fig:bootstrap_means_50}%
          \includegraphics[keepaspectratio=true, width=\linewidth]{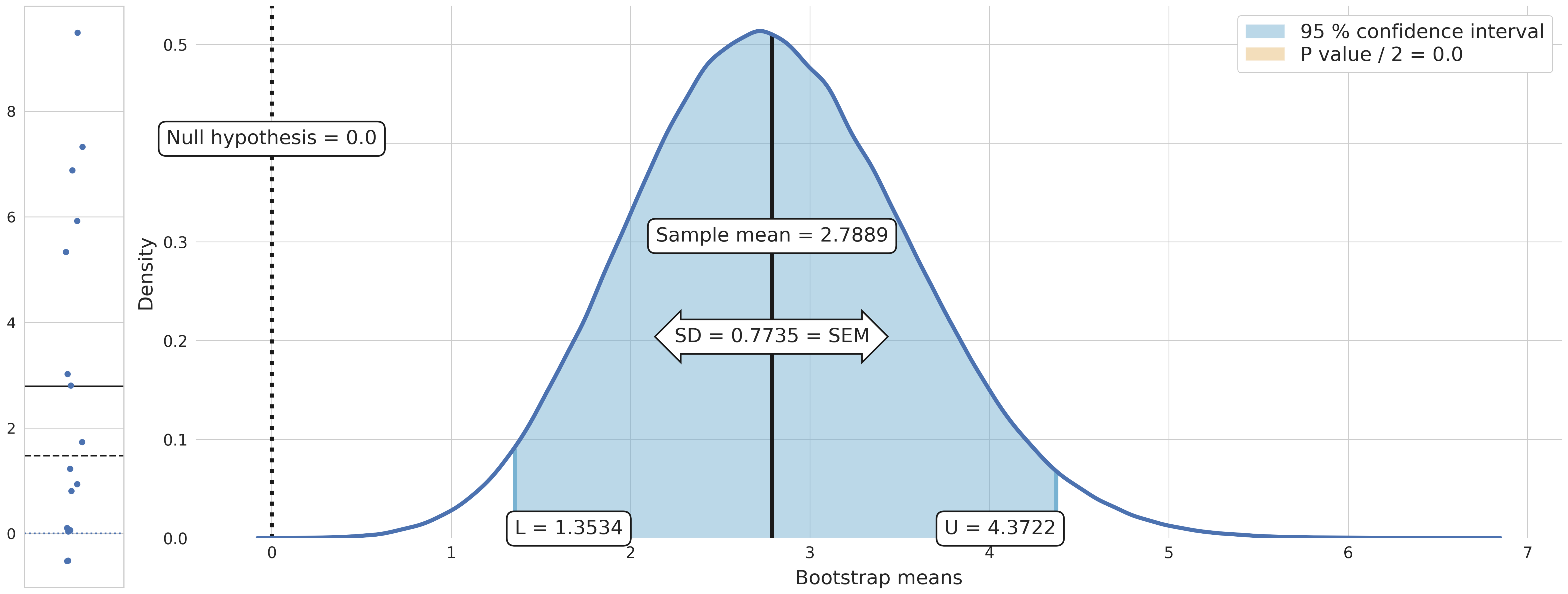}%
      }

      \subfloat[10~\% of the available training data]{%
		  \label{fig:bootstrap_means_10}%
          \includegraphics[keepaspectratio=true, width=\linewidth]{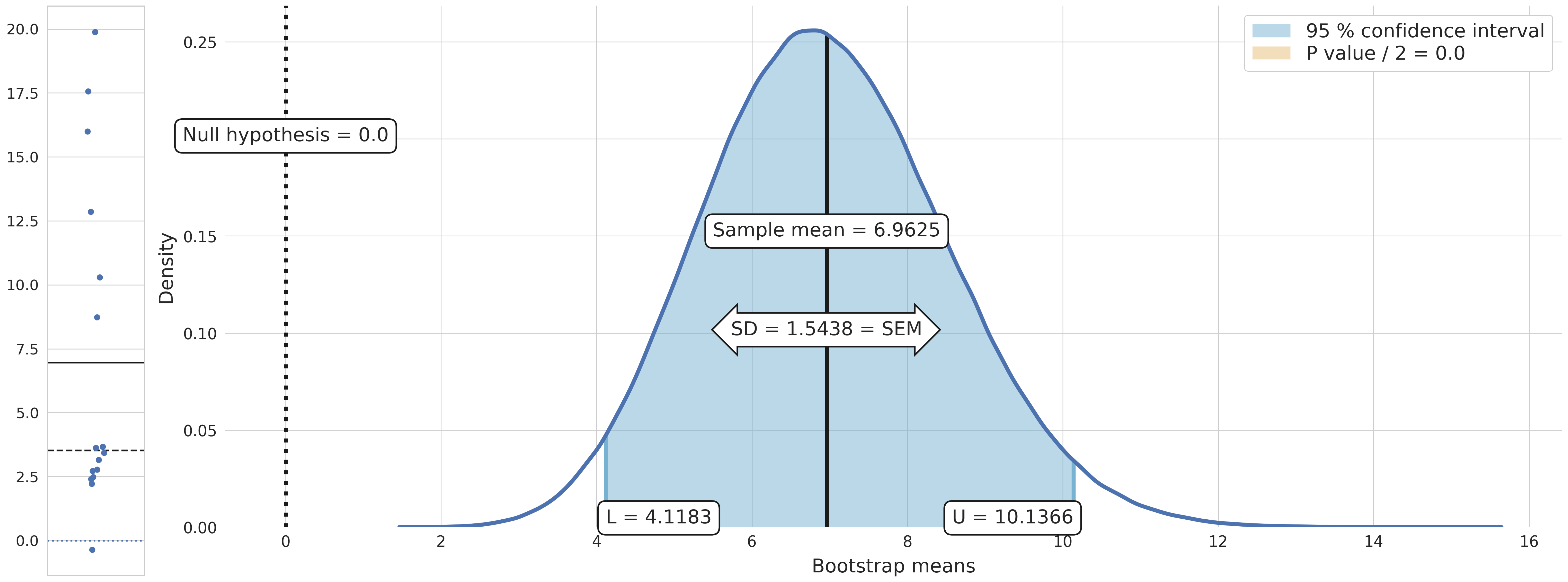}%
      }
  \caption{Bootstrap analysis analogous to the one detailed in \cref{sec:orig_nets} and \cref{fig:bootstrap_orig}, to analyze the statistical significance of the performance difference of models trained with only 10 and 50~\% of the data.}
  \label{fig:bootstrap_less_data}
\end{figure}

We argue that one of the main drawbacks of explicit regularization techniques is their poor adaptability to changes in the conditions with which the hyperparameters were tuned. To test this hypothesis and contrast it with the adaptability of data augmentation, we extended the analysis by training the same networks with fewer examples. All models were trained with the same random subset of data and evaluated in the same test set as the previous experiments. In order to better visualize how well each technique resists the reduction of training data, in \cref{fig:less_data} we show the fraction of baseline accuracy achieved by each model when trained with 50~\% and 10~\% of the available data. In this case, the baseline is thus each corresponding model trained with the complete data set. \Cref{tab:less_data} summarizes the mean and standard deviation of each combination and \cref{fig:bootstrap_less_data} shows the result of the bootstrap analysis. An extended report of results, including additional experiments with 80~\% and 1~\% of the data, is provided in the supplementary material (\cref{sec:suppl-results_details}).

One of the main conclusions of this set of experiments is that if no data augmentation is applied, explicit regularization hardly resists the reduction of training data by itself. On average, with 50~\% of the available data, these models only achieve 83.20~\% of the original accuracy (\cref{tab:less_data}), which, remarkably, is even worse than the models trained without any explicit regularization (88.11~\%). On 10~\% of the data, the average fraction is the same (58.75 and 58.72~\%, respectively). This implies that training with explicit regularization is even detrimental for the performance.

When combined with data augmentation, the models trained with explicit regularization (orange dots) also perform worse (88.78 and 61.16~\% with 50 and 10~\% of the data, respectively), than the models without explicit regularization (blue dots, 91.64 and 68.12~\% on average). Note that the difference becomes larger as the amount of available data decreases. Even more decisive are the results of the bootstrap analysis (\cref{fig:bootstrap_less_data}): the mean difference of the fraction of the performance achieved by the models trained without and with explicit regularization is 2.78 and 6.96, with 50 and 10~\% of the training data, respectively; the confidence intervals are well above the null hypothesis and the $P$ values are exactly 0.

Importantly, it seems that the combination of explicit regularization and data augmentation is only slightly better than training without data augmentation. Two reasons may explain this: first, the original regularization hyperparameters seem to adapt poorly to the new conditions. The hyperparameters were specifically tuned for the original setup and they would require re-tuning to obtain comparable results. Second, since explicit regularization reduces the representational capacity, this might prevent the models from taking advantage of the augmented data.

In contrast, the models trained without explicit regularization though with data augmentation more naturally adapt to the reduced availability of data. With 50~\% of the data, these models achieve about 91.5~\% of the performance with respect to training with the complete data sets. With only 10~\% of the data, they achieve nearly 70~\% of the baseline performance, on average. This highlights the suitability of data augmentation to serve, to a great extent, as true, useful data \cite{vinyals2016oneshot}.

\subsection{Shallower and deeper architectures}
\label{sec:depth}

\begin{figure}[htbp]
  \begin{center}
    \includegraphics[width = \linewidth]{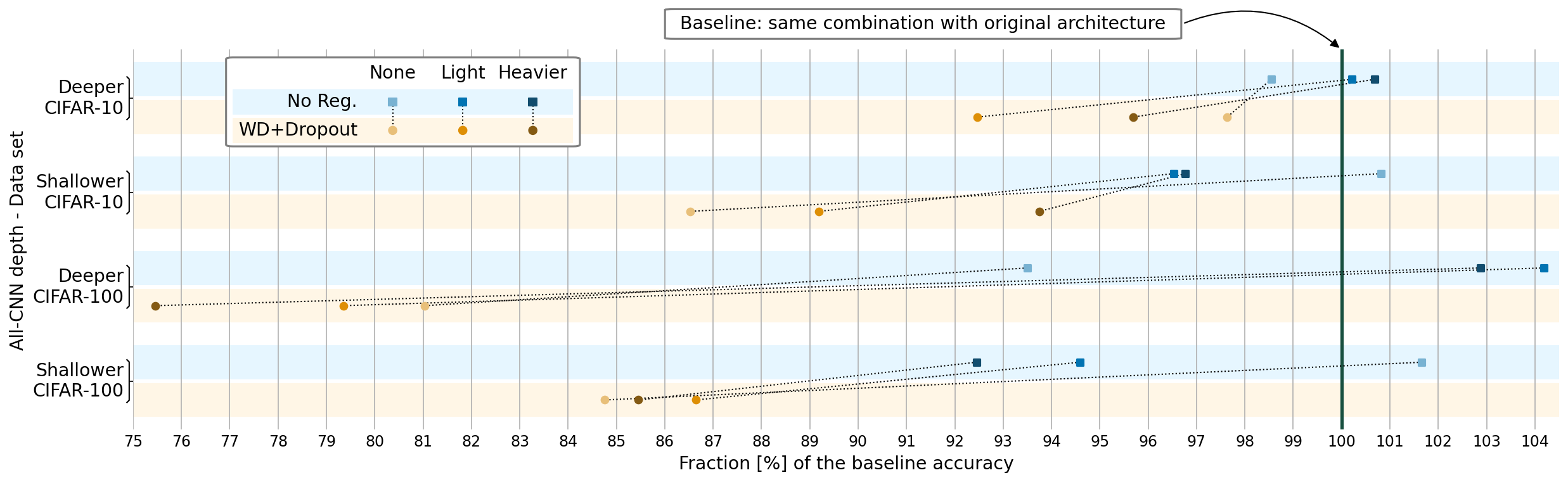}
  \end{center}
  \caption{Fraction of the original performance when the depth of the All-CNN architecture is increased or reduced in 3 layers. In the explicitly regularized models, the change of architecture implies a dramatic drop in the performance, while the models trained without explicit regularization present only slight variations with respect to the original architecture.}
  \label{fig:depth}
\end{figure}

\begin{figure}[htbp]
  \begin{center}
    \includegraphics[width = \textwidth]{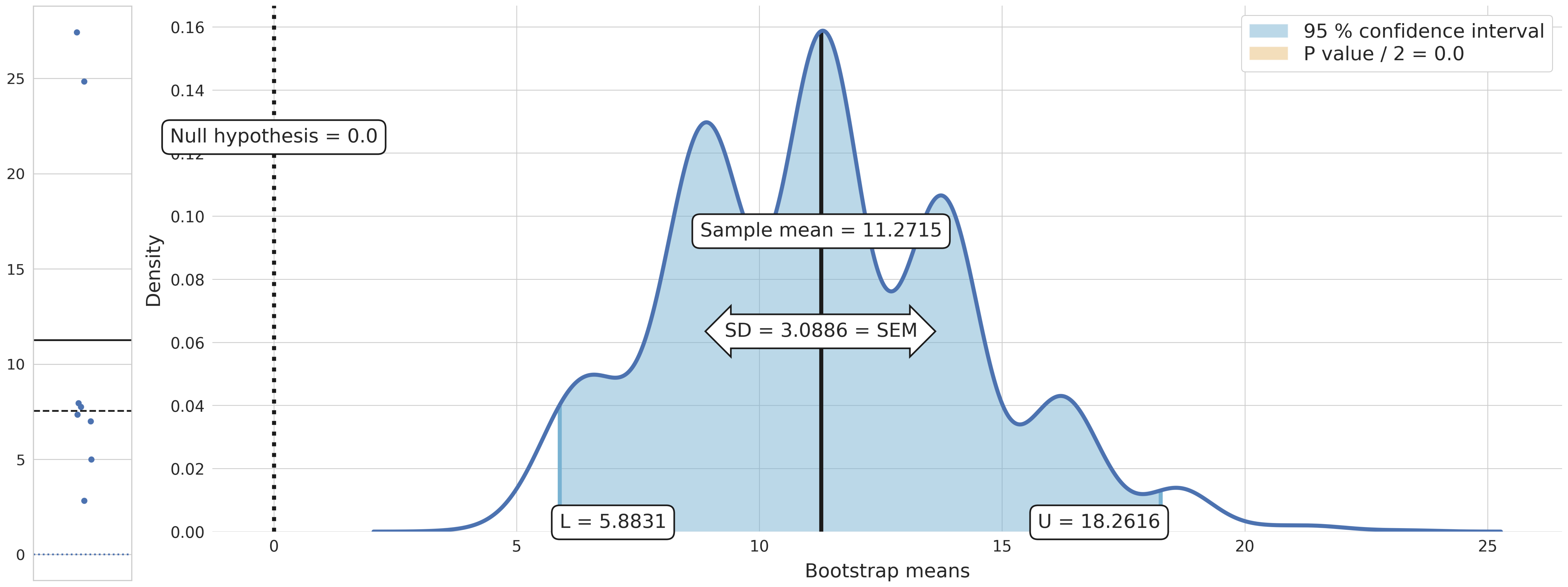}
  \end{center}
  \caption{Bootstrap analysis analogous to the one detailed in \cref{sec:orig_nets} and \cref{fig:bootstrap_orig}, to analyze the statistical significance of the performance difference of All-CNN trained with 3 more and 3 fewer layers.}
  \label{fig:bootstrap_depth}
\end{figure}

Finally, in the same spirit, we tested the adaptability of data augmentation and explicit regularization to changes in the depth of the All-CNN architecture, by training shallower (9 layers) and deeper (15 layers) versions of the architecture. We show the fraction of the performance with respect to the original architecture in \cref{fig:depth} and the bootstrap analysis in \cref{fig:bootstrap_depth}.

A noticeable result from these experiments is that all the models trained with weight decay and dropout (round orange dots) suffered a dramatic drop in performance when the architecture changed, regardless of whether deeper or shallower and of the amount of data augmentation. As a matter of fact, the models trained without explicit regularization performed on average 11.23~\% ($SD = 3.06$) better. As in the case of reduced training data, this may be explained by the poor adaptability of the regularization hyperparameters, which strongly depend on the architecture.

This highly contrasts with the performance of the models trained without explicit regularization (top, squared blue dots). With a deeper architecture, these models achieve slightly better performance, effectively exploiting the increased capacity. With a shallower architecture, they achieve only slightly worse performance\footnote{Note that the shallower models trained with neither explicit regularization nor data augmentation achieve even better accuracy than their counterpart with the original architecture, probably due to the reduction of overfitting provided by the reduced capacity.}. Thus, these models seem to more naturally adapt to the new architecture and data augmentation becomes beneficial.

It is worth commenting on the particular case of the CIFAR-100 benchmark, where the difference between the models with and without explicit regularization is even more pronounced, in general. It is common practice in object recognition papers to tune the parameters for CIFAR-10 and then test the performance on CIFAR-100 with the same hyperparameters. Therefore, these are typically less suitable for CIFAR-100. We believe this is the reason why the benefits of data augmentation seem even more pronounced on CIFAR-100 in our experiments.

In sum, these results highlight another crucial advantage of data augmentation: the effectiveness of its hyperparameters, that is the type of image transformations, depend mostly on the type of data, rather than on the particular architecture or amount of available training data, unlike explicit regularization hyperparameters, which require specific fine-tuning. Therefore, removing explicit regularization and training with data augmentation increases the flexibility of the models and their computational efficiency, hence reducing the environmental impact \cite{schwartz2019greenai}.

\section{Discussion}
\label{sec:discussion}

In this section we summarize our findings and discuss their relevance. In particular, we challenge the need for weight decay and dropout to train artificial neural networks, and propose to rethink data augmentation as a \textit{first class} technique instead of a \textit{cheating} method.

As an empirical analysis, one caveat of our work is the limited number of experiments (over 300 models trained). In order to increase the generality of our conclusions, we chose three significantly distinct network architectures and three data sets. Importantly, we also took a conservative approach in our experimentation: all the hyperparameters were kept as in the original models, which included both weight decay and dropout, as well as light augmentation. This setup is clearly sub-optimal for models trained without explicit regularization. Besides, the heavier data augmentation scheme was deliberately not optimized to improve the performance as it was not the scope of this work to propose a specific data augmentation technique. We leave for future work to explore data augmentation schemes that can more successfully be exploited by any deep model. Finally, in order to strengthen the conclusions from the empirical analysis, we have also discussed some theoretical insights in \cref{sec:theoretical}, concluding that the generalization gain provided by weight decay can be seen as a lower bound of what can be achieved by domain-specific data augmentation. We also hope that this work inspires researchers in other application domains, such as natural language processing, to further contrast data augmentation and explicit regularization.

\subsection{Do deep nets really need weight decay and dropout?}
\label{sec:wd_drop}
In \cref{sec:results} we have presented the results of a systematic empirical analysis of the role of weight decay, dropout and data augmentation in deep convolutional neural networks for object recognition. Our results have shown that explicit regularization is not only unnecessary \cite{zhang2016understandingdl}, but also that its performance gain can be achieved by data augmentation alone: in most cases, training with data augmentation only was better than training with both data augmentation and explicit regularization. In the few cases where that was not the case, the difference was very small. Moreover, unlike data augmentation, models trained with weight decay and dropout exhibited poor adaptability to changes in the architecture and the amount of training data. Why do researchers and practitioners keep training their neural networks with weight decay and dropout? Do deep nets really need weight decay and dropout?

The relevance of these findings lies in the fact that weight decay and dropout are almost ubiquitously present in artificial neural networks \cite{huang2017densenet, zagoruyko2016wrn, springenberg2014allcnn}, including recent, state of the art models \cite{tan2019efficientnet}. Certainly, it has repeatedly been shown that weight decay and dropout can boost the performance of neural networks, and we here do not challenge the usefulness of weight decay and dropout, but rather the convenience to use it, given the associated cost and risk, and available alternatives. 

First, not only do add weight decay and dropout extra computations during training, but also they typically require training the models several times with different hyperparameters: the coefficient of the penalty for weight decay; for dropout, the location of the dropout mask and the amount of units to drop. These hyperparameters are arguably very sensitive to changes in elements of the learning process. Here we have studied changes in the amount of training data (\cref{sec:less_data}) and the depth of the architecture (\cref{sec:depth}). 

Consider, for instance, our results in \cref{sec:less_data}: All-CNN trained on CIFAR-10 with weight decay, dropout and light augmentation reaches about 92~\% accuracy. If we were in the development process and were unsure about what architecture to use, we could simply try our network with three more layers and we would obtain about 85~\% accuracy. We could also try an architecture with three fewer layers and obtain about 82~\% accuracy. This may lead us to conclude that the first architecture tested has the right number of layers, because adding or removing layers drastically reduces the performance; or perhaps that by adding or removing layers there is some negative interaction between the layers sizes, or any other of the many hypothesis we could think of.

Consider now what happens if we train without weight decay and dropout: All-CNN trained on CIFAR-10 with light augmentation---but without weight decay and dropout---obtains about 93.3~\% accuracy. This is slightly better than the explicitly regularized model, but we will ignore this now. If we train this model with three more layers, we obtain 93.4~\% accuracy, that is the same or slightly better---as opposed to the drop of 7 points we have seen before. If we train with three fewer layers, we obtain about 90~\% accuracy, a drop of 3 points---as opposed to a drop of 10 points. In this case, we would not conclude that adding or removing layers creates negative interactions. Note that the only difference between these two cases is that the first models are trained with weight decay and dropout. Therefore, it may be reasonable to only include explicit regularization in the final version of a model, in order to potentially obtain a slight boost in performance prior to publication or production---provided the hyperparameters are adequately fine-tuned. Otherwise, keeping weight decay and dropout as intrinsic part of our models can certainly lead us astray.

Finally, we can draw some connections between these results from our empirical analysis and the theoretical insights from \cref{sec:theoretical} and the definitions of explicit and implicit regularization we have provided in \cref{sec:expl_impl_reg}. We have discussed that the role of explicit regularization techniques, such as weight decay and dropout, is to reduce the representational capacity of the models. This, according to statistical learning theory, can reduce overfitting and in turn improve generalization. However, artificial neural networks have usually orders of magnitude more parameters than training examples, and they still generalize well. While this phenomenon is yet to be fully understood, one intuitive hypothesis is that over-parameterization does not cause negative overfitting, but rather smooth fitting that can be suitable for accurate interpolation \cite{belkin2019biasvariance, hasson2020directfit}. Other recent articles have shed light into why over-parameterization is not a problem for generalization, and even a favorable regime \cite{brutzkus2017overparam, soltanolkotabi2018overparam, goldt2019overparam}. Hence, if over-parameterization is not a problem for artificial neural networks, is it then necessary to constrain the representational capacity through explicit regularization? Is it reasonable to train very large models, that require a lot of memory and computation---and negatively impact the environment---and at the same time constrain their capacity?

Relatedly, we hypothesize that a reason why artificial neural networks generalize well in many tasks in spite of (or thanks to) over-parameterization is the fact that the models include many sources of implicit regularization or, in other words, inductive biases. For example, it is known that stochastic gradient descent naturally converges to solutions with small norm \cite{zhang2016understandingdl, neyshabur2014implicitreg}, batch normalization also contributes to better generalization, convolutional layers are particularly efficient to process image data---and not only---to name a few examples. In our case, we argue that data augmentation has the potential to encode very powerful inductive biases that improve generalization. We conclude that in the presence of many other sources of implicit regularization and more effective inductive biases, weight decay and dropout may not be necessary to train large deep artificial neural networks\footnote{Previous work has suggested interesting connections between weight decay and other types of regularization and improved adversarial robustness \cite{galloway2018wdadversarial, jakubovitz2018regadversarial}. An interesting avenue for future work is studying whether this effects are also provided by data augmentation}. 

\subsection{Rethinking Data Augmentation}
\label{sec:rethink_daug}
Data augmentation is often regarded by authors of machine learning papers as \textit{cheating}, suggesting it should not be used in order to test the potential of newly proposed methods \cite{goodfellow2013maxout, graham2014fracmaxpool, larsson2016fractalnet}. In contrast, weight decay and dropout are considered intrinsic elements of the algorithms \cite{tan2019efficientnet}. In view of our results, we propose to rethink data augmentation and switch roles with explicit regularization: good models should effectively exploit data augmentation and explicit regularization should only be applied, if at all, once all other elements are fixed. This approach of training with data augmentation instead of explicit regularization improves the performance and saves valuable computational resources that are responsible for significant greenhouse gas emissions \cite{schwartz2019greenai, lannelongue2020carbonemissions}.

In this regard, it is worth highlighting some advantages of perceptually plausible data augmentation: Not only does it not reduce the representational capacity, unlike explicit regularization, but also, since the transformations reflect plausible variations of the real objects, it increases the robustness of the model \cite{novak2018sensitivity, rusak2020robustness}. As a matter of fact, data augmentation can be regarded as a way to exploit access to an oracle of the target function, since it directly introduces prior knowledge about the properties of human (visual) perception. Interestingly, it has also been shown that models trained with heavier data augmentation learn representations more aligned with the inferior temporal (IT) cortex, highlighting its connection with visual perception and biological vision \cite{hergar2018daugit}. 

Deep networks are especially well suited for data augmentation because they do not rely on pre-computed features. Moreover, unlike explicit regularization, it can be performed on the CPU, in parallel to the gradient updates. Finally, from \cref{sec:less_data,sec:depth} we concluded that data augmentation naturally adapts to architectures of different depth and amounts of available training data, without the need for specific fine-tuning of hyperparameters.

A commonly cited disadvantage of data augmentation is that it requires expert knowledge and is domain-specific \cite{devries2017daugfeatspace}. However, we argue that expert and domain knowledge should not be disregarded but exploited, since they constitute, in fact, useful inductive biases. A remarkable advantage of data augmentation is that a single augmentation scheme can be designed for a broad family of data---for example, natural images, using our knowledge about visual perception---and effectively applied to a broad set of tasks---object recognition, segmentation, localization, etc. We hope these insights encourage more research on data augmentation and, in general, highlight the importance of using the available data more effectively.

\section*{Acknowledgments}
This project has received funding from the European Union's Horizon 2020 research and innovation programme under the Marie Sklodowska-Curie grant agreement No 641805. Supported by BMBF and Max Planck Society.

We would like to thank Peter L. Bartlett for his expert feedback on the theoretical insights based on statistical learning (Section~\ref{sec:theoretical}).

\bibliographystyle{bibliography}
\bibliography{references}

%
%
%
%

\clearpage
\appendix

{
\section{Details of network architectures}
\label{sec:suppl-architectures}

This appendix presents the details of the network architectures used in the main experiments (\cref{sec:methods}): All-CNN, Wide Residual Network (WRN) and DenseNet. All-CNN is a relatively simple, small network with a few number of layers and parameters; WRN is deeper, has residual connections and many more parameters; and DenseNet is densely connected and is much deeper, but more parameter effective. 

\subsection{All Convolutional Network}
All-CNN consists exclusively of convolutional layers with ReLU activation \cite{glorot2011relu}, it is relatively shallow and has few parameters. For ImageNet, the network has 16 layers and 9.4 million parameters; for CIFAR, it has 12 layers and about 1.3 million parameters. In our experiments to compare the adaptability of data augmentation and explicit regularization to changes in the architecture (\cref{sec:depth}), we also test a \textit{shallower} version, with 9 layers and 374,000 parameters, and a \textit{deeper} version, with 15 layers and 2.4 million parameters. The four architectures can be described as in \cref{tab:suppl-allcnn}, where $K$\textbf{C}$D$($S$) is a $D \times D$ convolutional layer with $K$ channels and stride $S$, followed by batch normalization and a ReLU non-linearity. \textit{N.Cl.} is the number of classes and Gl.Avg. refers to global average pooling. The CIFAR network is identical to the All-CNN-C architecture in the original paper, except for the introduction of the batch normalization layers. The ImageNet version also includes batch normalization layers and a stride of 2 instead of 4 in the first layer to compensate for the reduced input size. 

\begin{table}[ht]
\caption{Specification of the All-CNN architectures.}
\begin{center}
\begin{tabular}{l|l}
\multirow{1}{*}{ImageNet}  & 96\textbf{C}11(2)--96\textbf{C}1(1)--96\textbf{C}3(2)--256\textbf{C}5(1)--256\textbf{C}1(1)--256\textbf{C}3(2)--384\textbf{C}3(1) \\
                           & --384\textbf{C}1(1)--384\textbf{C}3(2)--1024\textbf{C}3(1)--1024\textbf{C}1(1)--\textit{N.Cl}.C1(1)--Gl.Avg.--Softmax \\ [3pt]
\multirow{1}{*}{CIFAR}     & ~2$\times$96\textbf{C}3(1)--96\textbf{C}3(2)--2$\times$192\textbf{C}3(1)--192\textbf{C}3(2)--192\textbf{C}3(1)--192\textbf{C}1(1) \\
                           & --\textit{N.Cl}.C1(1)--Gl.Avg.--Softmax \\ [3pt]
\multirow{1}{*}{Shallower} & ~2$\times$96\textbf{C}3(1)--96\textbf{C}3(2)--192\textbf{C}3(1)--192\textbf{C}1(1)--\textit{N.Cl}.C1(1)--Gl.Avg.--Softmax \\ [3pt]
\multirow{1}{*}{Deeper}    & ~2$\times$96\textbf{C}3(1)--96\textbf{C}3(2)--2$\times$192\textbf{C}3(1)--192\textbf{C}3(2)--2$\times$192\textbf{C}3(1)--192\textbf{C}3(2) \\
                           & --192\textbf{C}3(1)--192\textbf{C}1(1)--\textit{N.Cl}.C1(1)--Gl.Avg.--Softmax \\
\end{tabular}
\end{center}
\label{tab:suppl-allcnn}
\end{table}

Importantly, we keep the same training parameters as in the original paper in the cases they are reported. Specifically, the All-CNN networks are trained using stochastic gradient descent, with fixed Nesterov momentum 0.9, learning rate of 0.01 and decay factor of 0.1. The batch size for the experiments on ImageNet is 64 and we train during 25 epochs decaying the learning rate at epochs 10 and 20. On CIFAR, the batch size is 128, we train for 350 epochs and decay the learning rate at epochs 200, 250 and 300. The kernel parameters are initialized according to the Xavier uniform initialization \cite{glorot2010glorot}.

\subsection{Wide Residual Network}
WRN is a modification of ResNet \cite{he2016resnet} that achieves better performance with fewer layers, but more units per layer. Here we choose for our experiments the WRN-28-10 version (28 layers and about 36.5 M parameters), which is reported to achieve the best results on CIFAR. It has the following architecture:

\begin{center}
\centering
16\textbf{C}3(1)--4$\times$160\textbf{R}--4$\times$320\textbf{R}--4$\times$640\textbf{R}--BN--ReLU--Avg.(8)--FC--Softmax
\end{center}

where $K$\textbf{R} is a residual block with residual function  BN--ReLU--$K$\textbf{C}3(1)--BN--ReLU--$K$\textbf{C} 3(1). BN is batch normalization, Avg.(8) is spatial average pooling of size 8 and FC is a fully connected layer. On ImageNet, the stride of the first convolution is 2. The stride of the first convolution within the residual blocks is 1 except in the first block of the series of 4, where it is set to 2 in order to subsample the feature maps. 

Similarly, we keep the training parameters of the original paper: we train with SGD, with fixed Nesterov momentum 0.9 and learning rate of 0.1. On ImageNet, the learning rate is decayed by 0.2 at epochs 8 and 15 and we train for a total of 20 epochs with batch size 32. On CIFAR, we train with a batch size of 128 during 200 epochs and decay the learning rate at epochs 60, 120 and 160. The kernel parameters are initialized according to the He normal initialization \cite{he2015he}.

\subsection{DenseNet}

The main characteristic of DenseNet \cite{huang2017densenet} is that the architecture is arranged into blocks whose layers are connected to all the layers below, forming a dense graph of connections, which permits training very deep architectures with fewer parameters than, for instance, ResNet. Here, we use a network with bottleneck compression rate $\theta = 0.5$ (DenseNet-BC), growth rate $k = 12$ and 16 layers in each of the three blocks. The model has nearly 0.8 million parameters. The specific architecture can be descried as follows:

\begin{center}
\centering
2$\times k$\textbf{C}3(1)--DB(16)--TB--DB(16)--TB--DB(16)--BN--Gl.Avg.--FC--Softmax
\end{center}

where DB($c$) is a dense block, that is a concatenation of $c$ convolutional blocks. Each convolutional block is of a set of layers whose output is concatenated with the input to form the input of the next convolutional block. A convolutional block with bottleneck structure has the following layers:

\begin{center}
\centering
BN--ReLU--4$\times k$\textbf{C}1(1)--BN--ReLU--$k$\textbf{C}3(1)--Concat.
\end{center}

TB is a transition block, which downsamples the size of the feature maps, formed by the following layers:

\begin{center}
\centering
BN--ReLU--$k$\textbf{C}1(1)--Avg.(2).
\end{center}

Like with All-CNN and WRN, we keep the training hyperparameters of the original paper. On the CIFAR data sets, we train with SGD, with fixed Nesterov momentum 0.9 and learning rate of 0.1, decayed by 0.1 on epochs 150 and 200 and training for a total of 300 epochs. The batch size is 64 and the are initialized with He initialization.

\section{Details of the heavier data augmentation scheme}
\label{sec:suppl-daug_details}

In this appendix we present the details of the heavier data augmentation scheme, introduced in \cref{sec:data}:

\begin{itemize}
  \item Affine transformations: 
  \vspace{5pt} \\
    $
      \begin{bmatrix}
      x^\prime \\
      y^\prime \\
      1
      \end{bmatrix}
      = 
      \begin{bmatrix}
      f_h z_x \cos(\theta) & -z_y \sin(\theta + \phi) & t_x \\
      z_x \sin(\theta) & z_y \cos(\theta + \phi) & t_y \\
      0 & 0 & 1
    \end{bmatrix}
    \begin{bmatrix}
      x \\
      y \\
      1
      \end{bmatrix}
    $
  \item Contrast adjustment: $x^\prime = \gamma (x - \overline{x}) + \overline{x}$

  \item Brightness adjustment: $x^\prime = x + \delta$
\end{itemize}

\begin{table}[ht]
\caption{Description and range of possible values of the parameters used for the heavier augmentation. $B(p)$ denotes a Bernoulli distribution and $\mathcal{U}(a, b)$ a uniform distribution.}
\begin{center}
\begin{tabular}{cll}
Parameter          & Description            & Range                                            \\
\hline \\
$f_h$              & Horiz. flip            & $1 - 2 B(0.5)$                                            \\
$t_x$              & Horiz. translation     & $\mathcal{U}(-0.1, 0.1)$                                  \\
$t_y$              & Vert. translation      & $\mathcal{U}(-0.1, 0.1)$                                  \\
$z_x$              & Horiz. scale           & $\mathcal{U}(0.85, 1.15)$                                 \\
$z_y$              & Vert. scale            & $\mathcal{U}(0.85, 1.15)$                                 \\
$\theta$           & Rotation angle         & $\mathcal{U}(-22.5^\circ, 22.5^\circ)$  \\
$\phi$             & Shear angle            & $\mathcal{U}(-0.15, 0.15)$                                \\
$\gamma$           & Contrast               & $\mathcal{U}(0.5, 1.5)$                                   \\
$\delta$           & Brightness             & $\mathcal{U}(-0.25, 0.25)$                                
\end{tabular}
\end{center}
\label{tab:suppl-heavier_aug}
\end{table}

\section{Detailed and extended experimental results}
\label{sec:suppl-results_details}

This appendix details the results of the main experiments shown in \cref{fig:orig,fig:less_data,fig:depth} and provides the results of many other experiments not presented in the main paper, so as not to clutter the visualization. Some of these results are the top-1 accuracy on ImageNet, the results of the models trained with dropout, but without weight decay; and the results of training with 80~\% and 1~\% of the data. 

Additionally, for many experiments we also trained some models without batch normalization. These results are provided within brackets in the tables. Note that the original All-CNN architecture \cite{springenberg2014allcnn} did not include batch normalization. In the case of WRN, we removed all batch normalization layers except the top-most one, before the spatial average pooling, since otherwise many models would not converge.

Recall that in the main paper we reported the improvement of accuracy with respect to a baseline, or the fraction of accuracy with respect to a baseline. Here, in the supplementary material, we report the absolute test accuracy of each model, for completeness.

\begin{table}[ht]
\caption{Test accuracy of All-CNN, WRN and DenseNet comparing the performance with and without explicit regularizers and the different augmentation schemes. Results within brackets show the performance of the models without batch normalization}
\begin{center}
\begin{tabular}{ccccccc}
\multicolumn{1}{c}{Network} & \multicolumn{1}{c}{WD} & \multicolumn{1}{c}{Dropout} & \multicolumn{1}{c}{Aug.} & \multicolumn{1}{c}{CIFAR-10} & \multicolumn{1}{c}{CIFAR-100} & \multicolumn{1}{c}{Acc. ImageNet} \\ \hline
\multirow{9}{*}{All-CNN}    & yes                    & yes                         & none                     & 90.04 (88.35)                & 66.50 (60.54)                 & 58.09                             \\
                            & yes                    & yes                         & light                    & 93.26 (91.97)                & 70.85 (65.57)                 & 63.35                             \\
                            & yes                    & yes                         & heavier                  & 93.08 (92.44)                & 70.59 (68.62)                 & 60.15                             \\ \cline{2-7} 
                            & no                     & yes                         & none                     & 77.99 (87.59)                & 52.39 (60.96)                 & ---                               \\
                            & no                     & yes                         & light                    & 77.20 (92.01)                & 69.71 (68.01)                 & ---                               \\
                            & no                     & yes                         & heavier                  & 88.29 (92.18)                & 70.56 (68.40)                 & ---                               \\ \cline{2-7} 
                            & no                     & no                          & none                     & 84.53 (71.98)                & 57.99 (39.03)                 & 56.53                             \\
                            & no                     & no                          & light                    & 93.26 (90.10)                & 69.26 (63.00)                 & 63.79                             \\
                            & no                     & no                          & heavier                  & 93.55 (91.48)                & 71.25 (71.46)                 & 61.37                             \\ \hline
\multirow{9}{*}{WRN}        & yes                    & yes                         & none                     & 91.44 (89.30)                & 71.67 (67.42)                 & 54.67                             \\
                            & yes                    & yes                         & light                    & 95.01 (93.48)                & 77.58 (74.23)                 & 68.84                             \\
                            & yes                    & yes                         & heavier                  & 95.60 (94.38)                & 76.96 (74.79)                 & 66.82                             \\ \cline{2-7} 
                            & no                     & yes                         & none                     & 91.47 (89.38)                & 71.31 (66.85)                 & ---                               \\
                            & no                     & yes                         & light                    & 94.76 (93.52)                & 77.42 (74.62)                 & ---                               \\
                            & no                     & yes                         & heavier                  & 95.58 (94.52)                & 77.47 (73.96)                 & ---                               \\ \cline{2-7} 
                            & no                     & no                          & none                     & 89.56 (85.45)                & 68.16 (59.90)                 & 61.29                             \\
                            & no                     & no                          & light                    & 94.71 (93.69)                & 77.08 (75.27)                 & 69.80                             \\
                            & no                     & no                          & heavier                  & 95.47 (94.95)                & 77.30 (75.69)                 & 69.30                             \\ \hline   
\multirow{6}{*}{DenseNet}   & yes                    & yes                         & none                     & 93.03 (---)                  & 71.62 (---)                   & ---                               \\
                            & yes                    & yes                         & light                    & 95.38 (---)                  & 75.17 (---)                   & ---                               \\
                            & yes                    & yes                         & heavier                  & 95.76 (---)                  & 75.44 (---)                   & ---                               \\ \cline{2-7} 
                            & no                     & no                          & none                     & 89.32 (---)                  & 69.56 (---)                   & ---                               \\
                            & no                     & no                          & light                    & 94.67 (---)                  & 74.69 (---)                   & ---                               \\
                            & no                     & no                          & heavier                  & 94.85 (---)                  & 75.14 (---)                   & ---                               
\end{tabular}
\end{center}
\label{tab:suppl-results_reg}
\end{table}

An important observation from \cref{tab:suppl-results_reg} is that the interaction of weight decay and dropout is not always consistent, since in some cases better results can be obtained with both explicit regularizers active and in other cases, only dropout achieves better generalization. In contrast, the effect of data augmentation seems to be consistent: just some light augmentation achieves much better results than training only with the original data set and performing heavier augmentation almost always further improves the test accuracy, without the need for explicit regularization.

Not surprisingly, batch normalization also contributes to improve the generalization of All-CNN and it seems to combine well with data augmentation. On the contrary, when combined with explicit regularization the results are interestingly not consistent in the case of All-CNN: it seems to improve the generalization of the model trained with both weight decay and dropout, but it drastically reduces the performance with only dropout, in the case of CIFAR-10 and CIFAR-100 without augmentation. A probable explanation is, again, that the regularization hyperparameters would need to be readjusted with a change of the architecture.

Furthermore, it seems that the gap between the performance of the models trained with and without batch normalization is smaller when they are trained without explicit regularization and when they include heavier data augmentation. This can be observed in \cref{tab:suppl-results_reg}, as well as in \cref{tab:suppl-results_lessdata_allcnn}, which contains the results of the models trained with fewer examples. It is important to note as well the benefits of batch normalization for obtaining better results when training with fewer examples. However, it is surprising that there is only a small drop in the performance of WRN---95.47~\% to 94.95~\% without regularization--- from removing the batch normalization layers of the residual blocks, given that they were identified as key components of ResNet \cite{he2016resnet, zagoruyko2016wrn}.

\begin{table}[htbp]
\caption{Test accuracy of All-CNN and WRN when training with only 80~\%, 50~\%, 10~\% and 1~\% of the available examples. Results within brackets correspond to the models without batch normalization}
\begin{center}
\begin{tabular}{ccccccc|cccc}
\multicolumn{1}{c}{Network}                                 & \multicolumn{1}{c}{Reg.} & \multicolumn{1}{c}{Aug.} & \multicolumn{4}{c}{CIFAR-10} & \multicolumn{4}{c}{CIFAR-100}        \\ \hline
                                                            &                                &                          & 80~\% & 50~\% & 10~\% & 1~\%  & 80~\% & 50~\% & 10~\% & 1~\%  \\ \cline{4-11}
\multirow{6}{*}{\makecell{All-CNN\\(no BN)}}                & yes                            & none                     & 86.61 & 82.33 & 61.61 & 29.90 & 52.51 & 44.94 & 19.79 & 3.60  \\
                                                            & yes                            & light                    & 91.25 & 87.37 & 69.18 & 26.85 & 63.24 & 54.68 & 22.79 & 3.65  \\
                                                            & yes                            & heavier                  & 91.42 & 88.94 & 64.14 & 26.87 & 65.89 & 57.91 & 26.29 & 2.52  \\ \cline{2-11} 
                                                            & no                             & none                     & 75.00 & 69.46 & 41.07 & 35.68 & 35.95 & 31.81 & 17.55 & 5.51  \\
                                                            & no                             & light                    & 88.75 & 84.38 & 67.65 & 29.29 & 56.81 & 47.84 & 24.34 & 5.36  \\
                                                            & no                             & heavier                  & 90.55 & 87.44 & 70.64 & 33.72 & 63.57 & 55.27 & 26.31 & 3.57  \\ \hline   
\multirow{6}{*}{\makecell{All-CNN\\(BN)}}                   & yes                            & none                     & 89.41 & 85.88 & 67.19 & 27.53 & 63.93 & 58.24 & 33.77 & 9.16  \\
                                                            & yes                            & light                    & 92.20 & 90.30 & 76.03 & 37.18 & 67.63 & 61.03 & 38.51 & 9.64  \\
                                                            & yes                            & heavier                  & 92.83 & 90.09 & 78.69 & 42.73 & 68.01 & 63.25 & 38.34 & 9.14  \\ \cline{2-11} 
                                                            & no                             & none                     & 83.04 & 78.61 & 60.97 & 38.89 & 55.78 & 48.62 & 26.05 & 9.50  \\
                                                            & no                             & light                    & 92.25 & 90.21 & 78.29 & 44.35 & 69.05 & 62.83 & 37.84 & 9.87  \\
                                                            & no                             & heavier                  & 92.80 & 90.76 & 79.87 & 47.60 & 69.40 & 64.41 & 39.85 & 11.45 \\ \hline   
\multirow{6}{*}{WRN}                                        & yes                            & none                     & 90.27 & 86.96 & 70.73 & 33.45 & 70.41 & 63.60 & 34.11 & 7.47  \\
                                                            & yes                            & light                    & 94.07 & 92.65 & 76.00 & 34.13 & 75.66 & 70.83 & 36.65 & 7.50  \\
                                                            & yes                            & heavier                  & 94.57 & 92.86 & 78.10 & 41.02 & 75.51 & 70.33 & 38.93 & 8.37  \\ \cline{2-11} 
                                                            & no                             & none                     & 88.98 & 85.56 & 60.39 & 38.63 & 66.10 & 60.64 & 23.65 & 9.47  \\
                                                            & no                             & light                    & 93.97 & 91.87 & 79.19 & 43.84 & 75.07 & 69.97 & 39.24 & 9.91  \\
                                                            & no                             & heavier                  & 94.84 & 92.77 & 80.29 & 47.14 & 75.38 & 70.72 & 41.44 & 11.03 \\ \hline   
\multirow{6}{*}{DenseNet}                                   & yes                            & none                     & ---   & 82.33 & 65.55 & 41.87 & ---   & 50.80 & 27.40 & 8.26  \\
                                                            & yes                            & light                    & ---   & 85.86 & 73.15 & 48.12 & ---   & 55.65 & 27.98 & 7.90  \\
                                                            & yes                            & heavier                  & ---   & 86.05 & 74.88 & 51.39 & ---   & 57.65 & 32.93 & 22.37 \\ \cline{2-11} 
                                                            & no                             & none                     & ---   & 83.03 & 65.45 & 41.25 & ---   & 52.70 & 30.88 & 8.49  \\
                                                            & no                             & light                    & ---   & 88.08 & 75.18 & 41.85 & ---   & 62.38 & 37.40 & 17.0  \\
                                                            & no                             & heavier                  & ---   & 90.29 & 76.52 & 47.70 & ---   & 62.59 & 40.53 & 20.45   
\end{tabular}
\end{center}
\label{tab:suppl-results_lessdata_allcnn}
\end{table}

The results in \cref{tab:suppl-results_lessdata_allcnn} clearly support the conclusion presented in \cref{sec:less_data} data augmentation alone better resists the lack of training data compared to explicit regularizers. Already with 80~\% and 50~\% of the data better results are obtained in some cases, but the differences become much bigger when training with only 10~\% and 1~\% of the available data. It seems that explicit regularization prevents the model from both fitting the data and generalizing well, whereas data augmentation provides useful transformed examples. Interestingly, with only 1~\% of the data, even without data augmentation the models without explicit regularization perform better. However, note that with only 1~\% of the data the performance is prone to larger variability, so more experimentation would be necessary to draw strong conclusions.

\begin{table}[t]
\caption{Test accuracy of the shallower and deeper versions of All-CNN on CIFAR-10 and CIFAR-100. Results in parentheses show the difference with respect to the original model.}
\begin{center}
\begin{tabular}{cccccc}
\multicolumn{1}{c}{Expl. Reg.} & \multicolumn{1}{c}{Aug.} & \multicolumn{2}{c}{CIFAR-10} & \multicolumn{2}{c}{CIFAR-100} \\ \hline     
                               &                          & Shallower      & Deeper           & Shallower     & Deeper             \\ \cline{3-6}
yes                            & no                       & 76.45 (-13.59) & 86.26 (-3.78)    & 51.31 (-9.23) & 49.06 (-11.48)     \\            
yes                            & light                    & 82.02 (-11.24) & 85.04 (-8.22)    & 56.81 (-8.76) & 52.03 (-13.54)     \\            
yes                            & heavier                  & 86.66 (-6.42)  & 88.46 (-4.62)    & 58.64 (-9.98) & 51.78 (-16.84)     \\ \hline
no                             & no                       & 85.22 (+0.69)  & 83.30 (-1.23)    & 58.95 (+0.96) & 54.22 (-3.77)      \\            
no                             & light                    & 90.02 (-3.24)  & 93.46 (+0.20)    & 65.51 (-3.75) & 72.16 (+2.90)      \\            
no                             & heavier                  & 90.34 (-3.21)  & 94.19 (+0.64)    & 65.87 (-5.38) & 73.30 (+2.35)      \\            
\end{tabular}
\end{center}
\label{tab:suppl-results_depth}
\end{table}

The same effect can be observed in \cref{tab:suppl-results_depth}, where both the shallower and deeper versions of All-CNN perform much worse when trained with explicit regularization, even when trained without data augmentation. This is another piece of evidence that explicit regularization needs to be used very carefully, it requires a proper tuning of the hyperparameters and is not always beneficial.

\section{Norm of the weight matrix}
\label{sec:suppl-norm_weight_matrix}

In this appendix we provide the computations of the Frobenius norm of the weight matrices of the models trained with different levels of explicit regularization and data augmentation, as a rough estimation of the complexity of the learned models. \Cref{tab:suppl-norm_reg} shows the Frobenius norm of the weight matrices of the models trained with different levels of explicit regularization and data augmentation. The clearest conclusion is that heavier data augmentation seems to yield solutions with larger norm. This is always true except in some All-CNN models trained without batch normalization. Another observation is that, as expected, weight decay constrains the norm of the learned function. Besides, the models trained without batch normalization exhibit smaller differences between different levels of regularization and augmentation and, in the case of All-CNN, less consistency.

\begin{table}[ht]
\caption{Frobenius norm of the weight matrices learned by the networks All-CNN and WRN on CIFAR-10 and CIFAR-100, trained with and without explicit regularizers and the different augmentation schemes. Norms within brackets correspond to the models without batch normalization}
\begin{center}
\begin{tabular}{cccccccc}
\multicolumn{1}{c}{WD} & \multicolumn{1}{c}{Dropout} & \multicolumn{1}{c}{Aug.} & \multicolumn{2}{c}{Norm CIFAR-10} & \multicolumn{2}{c}{Norm CIFAR-100} \\ \hline
                       &                             &                                 & All-CNN     & WRN                 & All-CNN      & WRN                 \\ \cline{4-7}
yes                    & yes                         & no                              & 48.7 (64.9) & 101.4 (122.6)       & 76.5 (97.9)  & 134.8 (126.5)       \\
yes                    & yes                         & light                           & 52.7 (63.2) & 106.1 (123.9)       & 77.6 (86.8)  & 140.8 (129.3)       \\
yes                    & yes                         & heavier                         & 57.6 (62.8) & 119.3 (125.3)       & 78.1 (83.1)  & 164.2 (132.5)       \\ \hline 
no                     & yes                         & no                              & 52.4 (70.5) & 153.3 (122.5)       & 79.7 (103.3) & 185.1 (126.5)       \\
no                     & yes                         & light                           & 57.0 (67.9) & 160.6 (123.9)       & 83.6 (93.0)  & 199.0 (129.4)       \\
no                     & yes                         & heavier                         & 62.8 (67.5) & 175.1 (125.2)       & 84.0 (88.0)  & 225.4 (132.5)       \\ \hline 
no                     & no                          & no                              & 37.3 (63.7) & 139.0 (120.4)       & 47.6 (102.7) & 157.9 (122.0)       \\
no                     & no                          & light                           & 47.0 (69.5) & 153.6 (123.2)       & 80.0 (108.9) & 187.0 (127.2)       \\
no                     & no                          & heavier                         & 62.0 (71.7) & 170.4 (125.4)       & 91.7 (91.7)  & 217.6 (132.9)       
\end{tabular}
\end{center}
\label{tab:suppl-norm_reg}
\end{table}

One of the relevant results presented in this paper is the poor performance of the regularized models on the shallower and deeper versions of All-CNN, compared to the models without explicit regularization (see \cref{tab:suppl-results_depth}). One hypothesis is that the \textit{amount} of regularization is not properly adjusted through the hyperparameters. This could be reflected in the norm of the learned weights, shown in \cref{tab:suppl-norm_depth}. However, the norm alone does not seem to fully explain the large performance differences between the different models. Finding the exact reasons why the regularized models not able to generalize well might require a much thorough analysis and we leave it as future work.

\begin{table}[ht]
\caption{Frobenius norm of the weight matrices learned by the shallower and deeper versions of the All-CNN network on CIFAR-10 and CIFAR-100.}
\begin{center}
\begin{tabular}{cccccc}
\multicolumn{1}{c}{Explicit Reg.} & \multicolumn{1}{c}{Aug. scheme} & \multicolumn{2}{c}{Norm CIFAR-10} & \multicolumn{2}{c}{Norm CIFAR-100} \\ \hline     
                                  &                                 & Shallower & Deeper                & Shallower & Deeper                 \\ \cline{3-6}
yes                               & no                              & 47.9      & 62.3                  & 68.9      & 92.1                   \\            
yes                               & light                           & 49.7      & 66.5                  & 67.1      & 95.7                   \\            
yes                               & heavier                         & 51.9      & 71.5                  & 66.2      & 96.9                   \\ \hline
no                                & no                              & 34.8      & 45.4                  & 64.7      & 53.4                   \\            
no                                & light                           & 45.6      & 57.3                  & 68.8      & 77.3                   \\            
no                                & heavier                         & 53.1      & 70.7                  & 68.3      & 97.5                        
\end{tabular}
\end{center}
\label{tab:suppl-norm_depth}
\end{table}

\section{Carbon footprint of the computational experiments}
\label{sec:suppl-carbon_footprint}

Training artificial neural networks effectively on large, non-trivial data sets consumes a considerable amount of energy \cite{strubell2019energydl}. Crucially, the amount of compute of the largest models has been increasing exponentially during the last decade \cite{amodei2018energyai}. Therefore, the contribution of deep learning research to global warming and climate change should not be neglected \cite{schwartz2019greenai, lacoste2019carbonemissions, lannelongue2020carbonemissions}. As authors of a research article that required training multiple neural network models, we wish to be as transparent as possible about the environmental impact of our experimental study and in this section we will report calculations of the estimated carbon emissions associated with training our models for this paper and the details of how these are computed.

In \cref{tab:architectures} of \cref{sec:archs} we report the estimated carbon emissions associated with training each architecture on each data set, given the specific characteristics of our computing hardware. These estimations rely on the online calculator available at \href{http://www.green-algorithms.org/}{green-algorithms.org}, developed by \citet{lannelongue2020carbonemissions}. In order to estimate the carbon emissions of each model, we took into consideration the following information: all the models were trained in a local desktop computer, located in Germany, with a single graphic processing unit (GPU), model GTX 1080 Ti, with 11 GB of memory. We assumed full usage of the 11 GB of memory and of the processing core for all models---a conservative estimation. 

\begin{table}[htbp]
\caption{Summary of estimated carbon emissions associated to training the models for our experimental setup}
\begin{center}
\begin{tabular}{ccccccc}
Network                                       & Data set                                                            & Depth                                & \% Data & N. models    & Total h.        & Total CO2e      \\ \hline
\multirow{10}{*}{All-CNN}                     & \multirow{7}{*}{\makecell{CIFAR\\2.5 h\\0.29 CO2e}}                 & \multirow{5}{*}{\makecell{original}} & 100~\%  & 36           & 90              & 10.73           \\
                                              &                                                                     &                                      & 80~\%   & 24           & 48              & 5.72            \\
                                              &                                                                     &                                      & 50~\%   & 24           & 30              & 3.58            \\
                                              &                                                                     &                                      & 10~\%   & 24           & 6               & 0.72            \\
                                              &                                                                     &                                      & 1~\%    & 24           & 0.6             & 0.072           \\ \cline{3-7}
                                              &                                                                     & shallower                            & 100~\%  & 12           & 25.2            & 3.0             \\ \cline{3-7}
                                              &                                                                     & deeper                               & 100~\%  & 12           & 36              & 4.29            \\ \cline{2-7}
                                              & \multirow{3}{*}{\makecell{ImageNet\\35--45 h\\4.17--5.36 CO2e}}     & \multirow{3}{*}{\makecell{original}} & 100~\%  & 6            & 270             & 32.18           \\
                                              &                                                                     &                                      & 50~\%   & 6            & 135             & 16.09           \\
                                              &                                                                     &                                      & 10~\%   & 6            & 27              & 3.22            \\ \hline
\multirow{8}{*}{WRN}                          & \multirow{5}{*}{\makecell{CIFAR\\14--15 h\\1.66--1.78 CO2e}}        & \multirow{5}{*}{\makecell{original}} & 100~\%  & 36           & 540             & 64.35           \\
                                              &                                                                     &                                      & 80~\%   & 12           & 144             & 17.16           \\
                                              &                                                                     &                                      & 50~\%   & 12           & 90              & 10.73           \\
                                              &                                                                     &                                      & 10~\%   & 12           & 18              & 2.15            \\
                                              &                                                                     &                                      & 1~\%    & 12           & 1.8             & 0.21            \\ \cline{2-7}
                                              & \multirow{3}{*}{\makecell{ImageNet\\100--145 h\\11.91--17.27 CO2e}} & \multirow{3}{*}{\makecell{original}} & 100~\%  & 6            & 870             & 103.68          \\
                                              &                                                                     &                                      & 50~\%   & 6            & 435             & 51.84           \\
                                              &                                                                     &                                      & 10~\%   & 6            & 87              & 10.37           \\ \hline
\multirow{3}{*}{DenseNet}                     & \multirow{3}{*}{\makecell{CIFAR\\24--27 h\\2.86--3.21 CO2e}}        & \multirow{3}{*}{\makecell{original}} & 100~\%  & 12           & 324             & 38.61           \\
                                              &                                                                     &                                      & 50~\%   & 6            & 81              & 9.65            \\
                                              &                                                                     &                                      & 10~\%   & 6            & 16.2            & 1.93            \\
                                              &                                                                     &                                      & 1~\%    & 6            & 1.62            & 0.19            \\ \hline
\vspace{-5pt}\\
\multicolumn{4}{c}{\textbf{Total}}                                                                                                                                   & \textbf{306} & \textbf{3276.4} & \textbf{390.45} 
\end{tabular}
\end{center}
\label{tab:suppl-carbon_emissions}
\end{table}

As reported in \cref{tab:suppl-carbon_emissions}, the complete set of experiments carried out for this work needed a total of 3,276 GPU hours (136.5 days), which correspond to actual real time, since we had access to a single GPU. With our hardware, this corresponds, according to \cite{lannelongue2020carbonemissions}, to an estimate of 832.53 kWh or 390.45 carbon dioxide equivalent (CO2e). Carbon dioxide equivalent represents the equivalent CO2 that would have the same global warming impact than a mixture of gases. By way of comparison, 390.45 CO2e corresponds to 34.25 tree-years---the time taken by a mature tree to absorb the CO2---, 69~\% of a flight New York City--San Francisco or 2,231 km in a passenger car.

\section{On the taxonomy of regularization}
\label{sec:suppl-reg_taxonomy}

Although it is out of the scope of this paper to elaborate on the taxonomy of regularization techniques for deep neural networks, a central contribution of this work is providing definitions of explicit and implicit regularization (see \cref{sec:reg-definitions}), which have been used ambiguously in the literature before. It is therefore worth mentioning here some of the previous work that has addressed the topic of regularization taxonomy or proposed other related terms.

In their textbook, \citet{goodfellow2016dlbook} review some of the most common regularization techniques used to train deep neural networks, but do not discuss the concepts of explicit and implicit regularization. More recently, \citet{kukavcka2017regularization} provided an extensive review of regularization methods for deep learning. Although they mentioned the implicit regularization effect of techniques such as SGD, no further discussion of the concepts is provided. Nonetheless, they defined the category \textit{regularization via optimization}, which is somewhat related to implicit regularization. However, regularization via optimization is more specific than our definition; hence, methods such as data augmentation would not fall into that category.

Recently, \citet{guo2018mixup} provided a distinction between \textit{data-independent} and \textit{data-dependent} regularization. They define data-independent regularization as those techniques that impose certain constraint on the hypothesis set, thus constraining the optimization problem. Examples are weight decay and dropout. We believe this is closely related to our definition of explicit regularization. On the other hand, they define data-dependent regularization as those techniques that make assumptions on the hypothesis set with respect to the training data, as is the case of data augmentation. While we acknowledge the usefulness of such taxonomy, we argue that the division between data-independent and -dependent regularization leaves some ambiguity about other techniques, such as batch-normalization, which neither imposes an explicit constraint on the representational capacity nor on the training data. 

On the contrary, our distinction between explicit and implicit regularization aims at being complete, since implicit regularization refers to any regularization effect that does not come from explicit---or data-independent---techniques.

Finally, we argue it would be useful to distinguish between domain-specific, \textit{perceptually plausible} data augmentation and other kinds of data-dependent regularization. Data augmentation originally aimed at creating new examples that could be plausible transformations of the real-world objects. In other words, the augmented samples should be no different in nature than the available data. In statistical terms, they should belong to the same underlying probability distribution. In contrast, one can think of data manipulations that would not mimic any plausible transformation of the data, which still can improve generalization and thus fall into the category of data-dependent regularization (and implicit regularization). One example is mixup, which is the subject of study in \cite{guo2018mixup}.
}

\end{document}